\documentclass{article} %
\usepackage{iclr2022_conference,times}

\usepackage{amsmath,amsfonts,bm}

\def\eqref#1{equation~\ref{#1}}

\def\1{\bm{1}}

\DeclareMathAlphabet{\mathsfit}{\encodingdefault}{\sfdefault}{m}{sl}
\SetMathAlphabet{\mathsfit}{bold}{\encodingdefault}{\sfdefault}{bx}{n}

\usepackage{hyperref}
\usepackage{url}

\usepackage{booktabs} %

\usepackage{amsmath}
\usepackage{amssymb}
\usepackage{mathtools}
\usepackage{amsthm}

\usepackage{array}     %
\usepackage{arydshln}  %
\usepackage{bbding}
\usepackage{blindtext}
\usepackage{bm}        %
\usepackage{capt-of}   %
\usepackage{caption}
\usepackage{colortbl}  %
\usepackage{comment}   %
\usepackage{dsfont}    %
\usepackage{enumitem}  %
\usepackage{floatrow}
\usepackage{graphicx}  %
\usepackage{multirow}  %
\usepackage{subcaption}
\usepackage{tabularx}
\usepackage{url}
\usepackage{wrapfig}
\usepackage{xspace}    %

\usepackage{algorithm}
\usepackage[noend]{algpseudocode}
\usepackage{algorithmicx}
\algnewcommand\algorithmicinput{\textbf{Input:}}
\algnewcommand\Input{\item[\algorithmicinput]}
\algnewcommand\algorithmicoutput{\textbf{Output:}}
\algnewcommand\Output{\item[\algorithmicoutput]}

\setlist[itemize]{itemsep=0.025cm}
\setlist[itemize]{leftmargin=*}

\usepackage{float}
\floatstyle{plaintop}
\restylefloat{table}

\DeclareMathOperator*{\Embed}{Embed}
\DeclareMathOperator*{\RNN}{RNN}
\DeclareMathOperator*{\Transformer}{Transformer}
\DeclareMathOperator*{\Modulate}{Modulate}
\DeclareMathOperator*{\MultiHead}{MultiHead}
\DeclareMathOperator*{\Concat}{Concat}
\DeclareMathOperator*{\film}{FiLM}

\DeclareMathOperator*{\Softmax}{softmax}

\DeclareMathOperator*{\Dense}{Dense}

\newcommand{\codenet}{Project~CodeNet\xspace}

\newcommand{\NeighborsIn}{N_\text{in}}

\newcommand{\nexit}{n_\text{exit}}
\newcommand{\nerror}{n_\text{error}}

\newcommand{\Error}{\text{error}}
\newcommand{\NoError}{\text{no error}}

\newcommand{\exceptioncolorname}{magenta\xspace}
\definecolor{exceptioncolor}{RGB}{185, 60, 63}
\definecolor{airforceblue}{rgb}{0.36, 0.54, 0.66}

\newfloatcommand{capbtabbox}{table}[][\FBwidth]

\newcommand\cruleblack[3][black]{\begingroup
    \fboxsep=0pt\raisebox{\fboxrule}{%
    \fbox{%
    \textcolor{#1}{%
        \rule{\dimexpr #2-2\fboxrule\relax}{\dimexpr #3-2\fboxrule\relax}}%
    }%
}\endgroup}

\newcommand\DAVIDHIDDEN[1]{}

\newcommand{\code}[1]{\texttt{#1}}

\title{Static Prediction of Runtime Errors \\ by Learning to Execute Programs \\ with External Resource Descriptions}

\author{David Bieber \\
Google Research \\
\texttt{dbieber@google.com} \\
\And
Rishab Goel \\
Mila\\
\texttt{rgoel0112@gmail.com} \\
\And
Daniel Zheng \\
Google Research \\
\texttt{danielzheng@google.com} \\
\And
Hugo Larochelle \\
Google Research \\
\texttt{hugolarochelle@google.com} \\
\And
Daniel Tarlow \\
Google Research \\
\texttt{dtarlow@google.com}
}

\iclrfinalcopy %
\begin{document}

\maketitle
\begin{abstract}
The execution behavior of a program often depends on external resources, such as program inputs or file contents, and so cannot be run in isolation.
Nevertheless, software developers benefit from fast iteration loops where automated tools identify errors as early as possible, even before programs can be compiled and run.
This presents an interesting machine learning challenge: can we predict runtime errors in a ``static'' setting, where program execution is not possible?
Here, we introduce a real-world dataset and task for predicting runtime errors, which we show is difficult for generic models like Transformers.
We approach this task by developing an interpreter-inspired architecture with an inductive bias towards mimicking program executions,
  which models exception handling and ``learns to execute'' descriptions of the contents of external resources.
Surprisingly, we show that the model can also predict the location of the error, despite being trained only on labels indicating the presence/absence and kind of error.
In total, we present a practical and difficult-yet-approachable challenge problem related to learning program execution and we demonstrate promising new capabilities of interpreter-inspired machine learning models for code.
\end{abstract}

\section{Introduction}	

We investigate applying neural machine learning methods to the static analysis of source code
  for the prediction of runtime errors.
The execution behavior of a program is in general not fully defined by its source code in isolation,
  because programs often rely on external resources like inputs, the contents of files, or the network.
Nevertheless, software developers benefit from fast iteration loops where automated tools identify errors early,
  even when program execution is not yet an option.
Therefore we consider the following machine learning challenge:
  can we predict runtime errors in a ``static'' setting, where program execution is not possible?

Recent work has made considerable progress toward applying machine learning to learning for code tasks \citep{allamanis2018survey}.	
Large language models have begun showing promise in source code generation,	
  but have struggled on tasks that require reasoning about the execution behavior of programs	
  \citep{codex, austin2021program}, requiring detailed step-by-step supervision to make headway on the task \citep{scratchpads}.	
We introduce the runtime error prediction task as a challenge problem, and in line with these findings,	
  observe that it is a challenging task for generic models like Transformers.	

This runtime error prediction task, along with the corresponding dataset that we introduce,	
  is well suited as a challenge problem because it is difficult-yet-approachable,	
  has immediate real-world value for software developers,	
  requires novel modeling considerations that we hypothesize will be applicable to a range of learning for code tasks,	
  and with this work, now has a suitable large dataset of real-world code.	
The task is to learn from data to predict whether provided source code is liable to exhibit a runtime error when it is run;	
  even when static analysis cannot provide guarantees of an error in the code, patterns learned from data may point to likely errors.	
Our dataset, Python Runtime Errors, consists of 2.4 million Python 3 programs written by competitive programmers selected from \codenet \citep{project-codenet}.	
We have run all programs in a sandboxed environment on sample inputs to determine their error classes,	
  finding the programs exhibit 26 distinct error classes including ``no error''.	
Each program relies on an external resource, the stdin input stream, and we pair the programs with a natural language description of the behavior of the stream.	
We make the task and prepared dataset, along with all models considered in this work, available for the research community to facilitate reproduction of this work and further research\footnote{https://github.com/google-research/runtime-error-prediction}.
To make progress on this challenging task,	
  we identify a promising class of models from prior work, interpreter-inspired models, and we demonstrate they perform well on the task.	
Instruction Pointer Attention Graph Neural Network (IPA-GNN) \citep{ipagnn} models simulate the execution of a program, following its control flow structure, but operating in a continuous embedding space.
In the IPA-GNN, a single step (or layer) of the model corresponds to a single step of execution in a traditional interpreter -- the execution of a single line -- and to accommodate the complexity of execution of real-world programs, we consider modeling up to 174 steps.

We make a number of improvements to IPA-GNN: scaling up to handle real-world code, adding the ability to ``learn to execute'' descriptions of the contents of external resources, and extending the architecture to model exception handling.	
We perform evaluations comparing these interpreter-inspired architectures against Transformer, LSTM, and GGNN baselines.	
Results show that our combined improvements lead to a large increase in accuracy in predicting runtime errors and to interpretability that allows us to predict the location of errors even though the models are only trained on the presence or absence and class of error.

In total, we summarize our contributions as:
\begin{itemize}
\item We introduce the runtime error prediction task and Python Runtime Errors dataset, with runtime error labels for millions of competition Python programs.
\item We demonstrate for the first time that IPA-GNN architectures are practical for processing real world programs, scaling them to a depth of 174 layers, and finding them to outperform generic models on the task.
\item We demonstrate that external resource descriptions such as descriptions of
the contents of stdin can be leveraged to improve performance on the task across all model architectures.
\item We extend the IPA-GNN to model exception handling, resulting in the Exception IPA-GNN,
which we find can localize errors even when only
trained on error presence and kind,
not error location.
\end{itemize}

\section{Related Work}

\paragraph{Execution behavior for identifying and localizing errors in source code}
Program analysis is a rich family of techniques for detecting defects in programs, including static analyses which are performed without executing code \citep{livshits2005finding,xie2006static,find-bugs} and dynamic analyses which are performed at runtime \citep{klee,sen2005cute,godefroid2005dart}. Linters and type checkers are popular error detection tools that use static analysis.

There has been considerable recent interest in applying machine learning to identifying and localizing errors in source code \citep{allamanis2018survey}.
\citet{project-codenet} makes a large dataset of real world programs available, which we build on in constructing the Python Runtime Errors dataset.
Though dozens of tasks such as variable misuse detection and identifying ``simple stupid bugs'' \citep{plur, allamanis2018learning,sstubs,great-paper} have been considered,
  most do not require reasoning about execution behavior, or else do so on synthetic data \citep{learning-to-execute, ipagnn}, or using step-by-step trace supervision \citep{scratchpads}.
\citet{codex} and \citet{austin2021program} find program execution to be a challenging task even for large-scale multiple billion parameter models.
In contrast with the tasks so far considered, the task we introduce requires reasoning about the execution behavior of real programs.
Several works including \citet{vasic-localize-and-repair} and \citet{beep-localization} perform error localization in addition to identification, but they use localization supervision, whereas this work does not.

\paragraph{Interpreter-inspired models}
Several neural architectures draw inspiration from program interpreters \citep{ipagnn, neural-turing-machines, differentiable-forth-interpreter, gaunt2017differentiable, neural-gpu, universal-transformer, neural-programmer-interpreters, differentiable-neural-computers}. Our work is most similar to \citet{ipagnn} and \citet{differentiable-forth-interpreter}, focusing on how interpreters handle control flow and exception handling, rather than on memory allocation and function call stacks.

\section{Runtime Error Prediction}
\label{sec:runtime-error-predictions}

The goal in the \emph{runtime error prediction} task is to determine statically
  whether a program is liable to encounter a runtime error when it is run.
The programs will not be executable, as they depend on external resources which are not available.
Descriptions of the contents of these external resources are available,
  which makes reasoning about the execution behavior of the programs plausible,
  despite there being insufficient information available to perform the execution.
Examples of external resources include program inputs, file contents, and the network.

Other approaches to finding runtime errors
  include using mocks to write unit tests,
  and applying static analysis techniques to reason about the set of plausible executions.
By instead framing this task as a machine learning challenge,
  it becomes applicable to the broad set of programs that lack unit tests,
  and it admits analysis even when these static analysis techniques are not applicable,
  for example in Python and when constraints are available only as natural language.
The other benefit of the task is as a challenge for machine learning models trained on code.
While there are many datasets in the literature that test understanding of different aspects
of code, we believe this task fills a gap: it is large-scale (millions of examples),
it has real-world implications and presents a practical opportunity for improvement using ML-based approaches,
and it tests a combination of statistical reasoning and reasoning about program execution.

We treat the task as a classification task,
  with each error type as its own class, with ``no error'' as one additional class,
  and with each program having only a single target.

\begin{table}[t]
    \caption{Distribution of target classes in the Python Runtime Errors dataset. $\dagger$ denotes examples in the balanced test split.}
    \label{tab:error-frequencies}
    \vspace{0pt}
    \centering
    \begin{center}
    \begin{small}
        \begin{tabular}{r|r|r|r}
    \toprule
Target Class        &  Train \#  & Valid \# &  Test \#  \\
\midrule
No error            & 1881303    & 207162   & 
            \begin{tabular}{@{}r@{}}205343 / \\ 13289$^\dagger$\end{tabular}   \\
\midrule
AssertionError      & 47         & 4        &  8        \\
AttributeError      & 10026      & 509      &  1674     \\
EOFError            & 7676       & 727      &  797      \\
FileNotFoundError   & 259        & 37       &  22       \\
ImportError         & 7645       & 285      &  841      \\
IndentationError    & 10         & 0        &  12       \\
IndexError          & 7505       & 965      &  733      \\
KeyError            & 362        & 39       &  22       \\
MemoryError         & 8          & 7        &  1        \\
ModuleNotFoundError & 1876       & 186      &  110      \\
NameError           & 21540      & 2427     &  2422     \\
numpy.AxisError     & 20         & 2        &  3        \\
OSError             & 19         & 2        &  2        \\
OverflowError       & 62         & 6        &  11       \\
re.error            & 5          & 0        &  0        \\
RecursionError      & 2          & 0        &  1        \\
RuntimeError        & 24         & 5        &  3        \\
StopIteration       & 3          & 0        &  1        \\
SyntaxError         & 74         & 4        &  3        \\
TypeError           & 21414      & 2641     &  2603     \\
UnboundLocalError   & 8585       & 991      &  833      \\
ValueError          & 25087      & 3087     &  2828     \\
ZeroDivisionError   & 437        & 47       &  125      \\
Timeout             & 7816       & 1072     &  691      \\
Other               & 18         & 8        &  2        \\
\bottomrule
    \end{tabular}

    \end{small}
    \end{center}
\end{table}

\subsection{Dataset: Python Runtime Errors}

We construct the \emph{Python Runtime Errors} dataset
  using submissions to competitive programming problems from \codenet \citep{project-codenet}.
Since this is an application-motivated task, we are interested in methods that can scale to handle the real-world complexity of these human-authored programs.
\codenet contains 3.28 million Python submissions attempting to solve 3,119 distinct competitive programming problems.
Each submission is a single-file Python program
    that reads from stdin and writes to stdout.
We run each Python submission in a sandboxed environment on a single input
    to determine the ground truth runtime error class for that program.
Programs are restricted to one second of execution time,
    and are designated as having a timeout error if they exceed that limit.
We filter the submissions keeping only
    those written in Python 3,
    which are syntactically valid,
    which do not make calls to user-defined functions,
    and which do not exceed a threshold length of 512 tokens once tokenized.  %
This results in a dataset of 2.44 million submissions,
    each paired with one of 26 target classes.
The ``no error'' target is most common, accounting for 93.4\% of examples.
For examples with one of the other 25 error classes,
    we additionally note the line number at which the error occurs,
    which we use as the ground truth for the unsupervised localization task of Section~\ref{sec:rq3}.
A consequence of using only a single input for each program is the ground truth labels
    under-approximate the full set of errors liable to appear at runtime,
    and a future improvement on the dataset might run each program on a wider range of valid inputs.

We divide the problems into train, validation, and test splits at a ratio of 80:10:10.
All submissions to the same problem become part of the same split.
This reduces similarities between examples across splits
    that otherwise could arise from the presence of multiple similar submissions for the same problem.
Since there is a strong class imbalance in favor of the no error class,
    we also produce a balanced version of the test split by sampling the no error examples
    such that they comprise roughly 50\% of the test split.
We use this balanced test split for all evaluations.

We report the number of examples having each target class in each split in Table~\ref{tab:error-frequencies}.
We describe the full dataset generation and filtering process in greater detail in Appendix~\ref{app:dataset-details}.

\section{Approach: IPA-GNNs as Relaxations of Interpreters}

We make three modifications to the Instruction Pointer Attention Graph Neural Network (IPA-GNN) architecture.
These modifications scale the IPA-GNN to real-world code, allow it to incorporate external resource descriptions into its learned executions, and add support for modeling exception handling.
The IPA-GNN architecture is a continuous relaxation of ``Interpreter A'' defined by the pseudocode in Algorithm~\ref{alg:ipa-gnn-interpreter}, minus the {\color{exceptioncolor}\exceptioncolorname} text.
We frame these modifications in relation to specific lines of the algorithm:
scaling the IPA-GNN to real-world code (Section~\ref{sec:per-node-embeddings}) and incorporating external resource descriptions (Section~\ref{sec:resource-descriptions}) both pertain to interpreting and executing statement $x_p$ at Line~3,
    and modeling exception handling adds the {\color{exceptioncolor}\exceptioncolorname} text at lines 4-6 to yield ``Interpreter B'' (Section~\ref{sec:exception-handling}).
We showcase the behavior of both Interpreters A and B on a sample program in Figure~\ref{tab:sample-program-and-execution},
    and illustrate an execution of the same program by a continuous relaxation of Interpreter B alongside it.

\begin{algorithm}
    \caption{\quad Interpreter implemented by {\color{exceptioncolor}Exc.} IPA-GNN}
    \label{alg:ipa-gnn-interpreter}
    \begin{algorithmic}[1]
        \Input Program $x$
        \State $h \gets \varnothing; p \gets 0, t \gets 0$     \Comment{Initialize the interpreter.}
        \While{$p \notin \{\nexit{\color{exceptioncolor}, \nerror}\}$}
            \State $h \gets \text{Evaluate}(x_p, h)$  \Comment{Evaluate the current line.}
            { \color{exceptioncolor}
            \If{$\text{Raises}(x_p, h)$}
                \State $p \gets \text{GetRaiseNode}(x_p, h)$  \Comment{Raise exception.}
            \Else%
            }%
            \If{$\text{Branches}(x_p, h)$}
                \State $p \gets \text{GetBranchNode}(x_p, h)$  \Comment{Follow branch.}
            \Else
                \State $p \gets p + 1$  \Comment{Proceed to next line.}
            \EndIf
            \State $t \gets t + 1$
        \EndWhile
    \end{algorithmic}
\end{algorithm}

\begin{figure*}[h]
    \caption{A sample program and its execution under discrete interpreters A and B (Algorithm~\ref{alg:ipa-gnn-interpreter}) and under a continuous relaxation $\tilde{B}$ of Interpreter B.}
    \label{tab:sample-program-and-execution}
    \begin{subtable}[h]{0.25\textwidth}
        \centering
        \begin{tabular}{rl}
        \toprule
         $n$ &    \begin{sc}Source\end{sc} \\
        \midrule
           1 &      \code{x~=~input()} \\
           2 &        \code{if~x~>~0:} \\
           3 &    \code{~~y~=~4/3~*~x} \\
           4 &            \code{else:} \\
           5 &     \code{~~y~=~abs(x)} \\
           6 &  \code{z~=~y~+~sqrt(x)} \\
           7 &           \code{<exit>} \\
           8 &          \code{<raise>} \\
        \bottomrule
        \end{tabular}
       \caption{Sample program illustrative of Algorithm~\ref{alg:ipa-gnn-interpreter} behavior.}
       \label{tab:sample-source}
    \end{subtable}
    \hfill
    \begin{subtable}[h]{0.70\textwidth}
        \centering
                    \begin{subtable}[h]{\textwidth}
                    \centering
    \begin{sc}
    \begin{tabular}{rl}
    \toprule
    StdIn &  \code{-3}          \\
    StdIn Description & \code{"A single integer -10..10"} \hspace{.62cm}        \\
    \bottomrule
    \end{tabular}
    \end{sc}
     \caption{Resource description indicates likely values the sample program will receive on stdin.}
     \label{tab:stdin-description}
                \end{subtable}
                \vfill\vspace{2mm}
                \begin{subtable}[h]{\textwidth}
                    \centering
    \begin{tabular}{rlrrll}
    \toprule
    $t$ &                                        $h_{A,B}$ & $p_A$ & $p_B$ & $h_{\tilde{B}}$                     & $p_{\tilde{B}}$ \\
    \midrule
    0 &                                        \code{\{\}} &     1 &     1 & \cruleblack[white]{8pt}{8pt}        & \code{[10000000]} \\
    1 &                                   \code{\{x:~-3\}} &     2 &     2 & \cruleblack[airforceblue]{8pt}{8pt} & \code{[01000000]} \\
    2 &                                   \code{\{x:~-3\}} &     5 &     5 & \cruleblack[airforceblue]{8pt}{8pt} & \code{[00001000]} \\
    3 &                             \code{\{x:~-3, y:~3\}} &     6 &     6 & \cruleblack[blue]{8pt}{8pt}         & \code{[00000100]} \\
    4 &             \code{ValueError(lineno=6)} &     7 &     8 & \cruleblack[yellow]{8pt}{8pt}       & \code{[00000001]} \\
    \bottomrule
    \end{tabular}
                    \caption{Step-by-step execution of the sample program according to Interpreters A and B, and under continuous relaxation $\tilde{B}$. Distinct colors represent distinct embedding values.}
                    \label{tab:sample-execution}
                 \end{subtable}
     \end{subtable}
\end{figure*}
\setcounter{figure}{1}

\subsection{Extending the IPA-GNN to Real-world Code}
\label{sec:per-node-embeddings}

\citet{ipagnn} interprets the IPA-GNN architecture as a message passing graph neural network operating on the statement-level control flow graph of the input program $x$.
Each node in the graph corresponds to a single statement in the program.
At each step $t$ of the architecture, each node performs three steps:
it executes the statement at that node (Line 3, Equation~\ref{equ:execute}),
computes a branch decision (Lines 7-8, Equation~\ref{equ:branch-decision}),
and performs mean-field averaging over the resulting states and instruction pointers (Equations~5 and 6 in \citet{ipagnn}).

Unlike in \citet{ipagnn} where program statements are simple enough to be uniformly encoded as four-tuples,
    the programs in the Python Runtime Errors dataset consist of arbitrarily complex Python statements authored by real programmers in a competition setting.
The language features used are numerous and varied, and so the statement lengths vary substantially, with a mean statement length of 6.7 tokens;
    we report the full distribution of statement lengths in Figure~\ref{fig:statement-lengths}.

The IPA-GNN architecture operates on a program $x$'s statement-level control flow graph,
    and so requires per-statement embeddings $\Embed(x_n)$ for each statement $x_n$.
We first apply either a \emph{local} or \emph{global} Transformer encoder to produce per-token embeddings,
  and we subsequently apply one of four pooling variants to a span of such embeddings to produce a \emph{node embedding} per statement in a program.
In the local approach, we apply an attention mask to
  limit the embedding of a token in a statement to attending to other tokens in the same statement.
In the global approach, no such attention mask is applied, and so every token may attend to every other token in the program. %
We consider four types of pooling in our hyperparameter search space: \emph{first}, \emph{sum}, \emph{mean}, and \emph{max}. The resulting embedding is given by
\begin{equation}
    \label{equ:embed}
    \Embed(x_n) = \text{Pool}\!\left(\Transformer(x)_{\text{Span}(x, n)}\right).
\end{equation}
First pooling takes the embedding of the first token in the span of node $n$.
Sum, mean, and max pooling apply their respective operations to the embeddings of all tokens in the span of node $n$.

Finally we find that real-world code requires as many as 174 steps of the IPA-GNN under the model's heuristic for step limit $T(x)$ (Appendix~\ref{app:step-limit}).
To reduce the memory requirements, we apply rematerialization
at each step of the model \citep{gradient-checkpointing, deep-sublinear}.

\subsection{Executing with Resource Descriptions}
\label{sec:resource-descriptions}

Each program $x$ in the dataset is accompanied by a description of what values stdin may contain at runtime,
which we tokenize to produce embedding $d(x)$.
Analogous to Line~1 of Algorithm~\ref{alg:ipa-gnn-interpreter}, IPA-GNN architectures initialize with per-node hidden states $h_{0, :} = 0$ and soft instruction pointer $p_{0, n} = \mathds{1}\{n=0\}.$
Following initialization, each step of an IPA-GNN begins by simulating execution (Line 3) of each non-terminal statement with non-zero probability under the soft instruction pointer to propose a new hidden state contribution
\begin{equation}
    \label{equ:execute}
    a^{(1)}_{t, n} = \RNN(h_{t-1, n}, {\color{exceptioncolor}\Modulate(}\!\Embed(x_n){\color{exceptioncolor}, d(x), h_{t-1,n})}).%
\end{equation}
Here, the text in {\color{exceptioncolor}\exceptioncolorname} shows our modification to the IPA-GNN architecture to incorporate external resource descriptions.
We consider both \emph{Feature-wise Linear Modulation} (FiLM) \citep{film} and \emph{cross-attention} \citep{cross-attention} for the $\Modulate$ function, which we define in Appendix~\ref{app:modulation-function}.
This modulation allows the IPA-GNN to execute differently at each step depending on what information the resource description provides, whether that be type information, value ranges, or candidate values.

We also consider one additional method that applies to any model: injecting the description as a \emph{docstring} at the start of the program.
This method yields a new valid Python program, and so any model can accommodate it.

\subsection{Modeling Exception Handling}
\label{sec:exception-handling}

The final modification we make to the IPA-GNN architecture is to model exception handling.
In Algorithm~\ref{alg:ipa-gnn-interpreter}, this corresponds to adding the {\color{exceptioncolor}\exceptioncolorname} text to form Interpreter B, computing a raise decision (Lines 4-6, Equation~\ref{equ:raise-decision}).
We call the architecture that results the ``Exception IPA-GNN''.

Whereas execution always proceeds from statement to next statement in Interpreter~A and in the IPA-GNN, Interpreter~B admits another behavior. In Interpreter~B and the Exception IPA-GNN,
execution may proceed from any statement to a surrounding ``except block'', if it is contained in a try/except frame, or else to a special global error node, which we denote $\nerror$.
In the sample execution in Figure~\ref{tab:sample-program-and-execution}c we see at step $t=4$ the instruction pointer $p_B$ updates to $\nerror=8$.

We write that the IPA-GNN makes raise decisions as
\begin{equation}
    \label{equ:raise-decision}
    {\color{exceptioncolor}
    b_{t,n,r(n)}, (1 - b_{t,n,r(n)})
    =
    \Softmax
    \left( \Dense(a^{(1)}_{t,n}) \right)%
    }.
\end{equation}
The dense layer here has two outputs representing the case that an error is raised and that no error is raised.
Here $r(n)$ denotes the node that statement $n$ raises to in program $x$; $r(n) = \nerror$ if $n$ is not contained in a try/except frame, and $b_{t,n,n'}$ denotes the probability assigned by the model to transitioning from executing $n$ to $n'$.

Next the model makes soft branch decisions in an analogous manner; the dense layer for making branch decisions has distinct weights from the layer for making raise decisions.
\begin{equation}
    \label{equ:branch-decision}
    b_{t,n,n_1}, b_{t,n,n_2}
    =
    (1~{\color{exceptioncolor} -~b_{t,n,r(n)}}) \cdot \Softmax
    \left( \Dense(a^{(1)}_{t,n}) \right).
\end{equation}
The text in {\color{exceptioncolor}\exceptioncolorname} corresponds to the ``else'' at Line~6.
The model has now assigned probability to up to three possible outcomes for each node: the probability that $n$ raises an exception $b_{t,n,r(n)}$, the probability that the true branch is followed $b_{t,n,n_1}$, and the probability that the false branch is followed $b_{t,n,n_2}$.
In the common case where a node is not a control node and has only one successor, rather than separate true and false branches, the probability of reaching that successor is simply $1 - b_{t,n,r(n)}$.

Finally, %
we assign each program a step limit $T(x)$ using the same heuristic as \citet{ipagnn}, detailed in Appendix~\ref{app:step-limit}.
After $T(x)$ steps of the architecture, the model directly uses the probability mass at $\nexit$ and $\nerror$ to predict whether the program raises an error,
  and if so it predicts the error type using the hidden state at the error node.
We write the modified IPA-GNN's predictions as
\begin{align}
    {\color{exceptioncolor}P(\NoError)}~&{\color{exceptioncolor}\propto p_{T(x), \nexit}} \text{ and}\\
    {\color{exceptioncolor}P(\Error)}   ~&{\color{exceptioncolor}\propto p_{T(x), \nerror}}, \text{ with }\\
    {\color{exceptioncolor}P(\Error = k \mid \Error)}~ &{\color{exceptioncolor}=\Softmax\left(\Dense(h_{T(x), \nerror})\right)}.
\end{align}
We train with a cross-entropy loss on the class predictions, treating ``no error'' as its own class.

\subsection{Unsupervised Localization of Errors}

When we model exception handling explicitly in the IPA-GNN, the model makes soft decisions as to when to raise exceptions.
We can interpret these decisions as predictions of the location where a program might raise an error.
We can then evaluate how closely these location predictions match the true locations where exceptions are raised,
    despite never training the IPA-GNN with supervision that indicates error locations.

For programs that lack try/except frames, we compute the localization predictions of the model by summing,
    separately for each node, the contributions from that node to the exception node across all time steps.
This gives an estimate of \emph{exception provenance} as
\begin{equation}
    p(\text{error at statement }n) = \sum_{t} {p_{t,n} \cdot b_{t,n,\nerror}}.
\end{equation}
For programs with a try/except frame, we must trace the exception back to the statement that originally raised it.
We provide the calculation for this in Appendix~\ref{app:localization}.

\section{Experiments}

In our experiments we evaluate the following research questions:

\textbf{RQ1:} How does the adaptation of the IPA-GNN to real-world code compare against standard architectures like GGNN, LSTM, and Transformer? (Section~\ref{sec:rq1})

\textbf{RQ2:} What is the impact of including resource descriptions? What methods for incorporating them work best? (Section~\ref{sec:rq2})

\textbf{RQ3:} How interpretable are the latent instruction pointer quantities inside the Exception IPA-GNN for localizing where errors arise? How does unsupervised localization with the Exception IPA-GNN compare to alternative unsupervised approaches to localization based on multiple instance learning and standard architectures? (Section~\ref{sec:rq3})

\subsection{Evaluation of IPA-GNN Against Baselines}
\label{sec:rq1}

We describe our experimental setup for our first experiment, comparing the IPA-GNN architectures with
    Transformer \citep{transformer},
    GGNN \citep{ggnn},
    and LSTM \citep{lstm} baselines.
In all approaches, we use the 30,000 token vocabulary constructed in Appendix~\ref{app:dataset-details},
  applying Byte-Pair Encoding (BPE) tokenization \citep{bytepairencoding} to tokenize each program into a sequence of token indices.
The Transformer operates on this sequence of token indices directly, with its final representation computed via mean pooling.
For all other models (GGNN, LSTM, IPA-GNN, and Exception IPA-GNN), the token indices are first combined via a masked (local) Transformer to produce per-node embeddings,
  and the model operates on these per-node embeddings, as in Section~\ref{sec:per-node-embeddings}.
Following \citet{ipagnn} we encode programs for a GGNN using six edge types,
    and use a two-layer LSTM for both the LSTM baseline and for the RNN in all IPA-GNN variants.

For each approach, we perform an independent hyperparameter search using random search.
We list the hyperparameter space considered
    and model selection criteria
    in Appendix~\ref{app:baseline-hparams}.
The models are each trained
    to minimize a cross-entropy loss on the target class
    using stochastic gradient descent for up to 500,000 steps with a mini-batch size of 32.
In order to more closely match the target class distribution found in the balanced test set,
    we sample mini-batches such that the proportion of examples with target ``no error'' and those with an error target is 1:1 in expectation.
We evaluate the selected models on the balanced test set, and report the results in Table~\ref{tab:main-results} (see rows without check marks).
Weighted F1 score weights per-class F1 scores by class frequency, while weighted error F1 score restricts consideration to examples with a runtime error.

\begin{table}[h]
    \caption{Accuracy, weighted F1 score (W. F1), and weighted error F1 score (E. F1) on the Python Runtime Errors balanced test set.}
    \label{tab:main-results}
    \centering 
    \begin{center}
    \begin{small}
    \begin{sc}

\begin{tabular}{clclll}
\toprule
& Model &   R.D.? &  Acc. & W. F1 & E. F1 \\
\midrule
\multirow{3}{*}{\rotatebox[origin=c]{90}{\parbox[c]{1cm}{\centering Base-lines}}} &
GGNN        &                 &  62.8 &  58.9 &  45.8  \\
& Transformer &                 &  63.6 &  60.4 &  48.1  \\
& LSTM        &                 &  66.1 &  61.4 &  48.4  \\
\midrule
\multirow{5}{*}{\rotatebox[origin=c]{90}{\parbox[c]{1.8cm}{\centering Ablations}}} &
GGNN        & \CheckmarkBold  &  68.3 &  66.5 &  56.8 \\
& Transformer & \CheckmarkBold  &  67.3 &  65.1 &  54.7 \\
& LSTM        & \CheckmarkBold  &  68.1 &  66.8 &  58.3 \\
& IPA-GNN     &                 &  68.3 &  64.8 &  53.8  \\
& E. IPA-GNN  &                 &  68.7 &  64.9 &  53.3  \\
\midrule
\multirow{2}{*}{\rotatebox[origin=c]{90}{\parbox[c]{.7cm}{\centering Ours}}} &
IPA-GNN     &  \CheckmarkBold & 71.4 &  70.1 &  62.2 \\
& E. IPA-GNN  &  \CheckmarkBold & \textbf{71.6} &  \textbf{70.9} &  \textbf{63.5} \\
\bottomrule
\end{tabular}
    \end{sc}
    \end{small}
    \end{center}
\end{table}

We perform additional evaluations using the same experimental setup but distinct initializations to compute measures of variance, which we detail in Appendix~\ref{app:variance}.

\textbf{RQ1:} The interpreter-inspired architectures show significant gains over the GGNN, Transformer, and baseline approaches on the runtime error prediction task.
We attribute this to the model's inductive bias toward mimicking program execution.

\subsection{Incorporating Resource Descriptions}
\label{sec:rq2}

We next evaluate methods of incorporating resource descriptions into the models.
For each architecture we apply the docstring approach of processing resource descriptions of Section~\ref{sec:resource-descriptions}.
This completes a matrix of ablations, allowing us to distinguish the effects due to architecture change from the effect of the resource description.
We follow the same experimental setup as in Section~\ref{sec:rq1}, and
  show the results in Table~\ref{tab:main-results} (compare rows with check marks to those without).

We also consider the FiLM and cross-attention methods of incorporating resource descriptions into the IPA-GNN.
Following the same experimental setup again,
  we show the results of this experiment in Table~\ref{tab:fusion-results}.
Note that the best model overall by our model selection criteria on validation data was the IPA-GNN with cross-attention, though the Exception IPA-GNN performed better on test.

\begin{table*}[t]
    \caption{A comparison of early and late fusion methods for incorporating external resource description information into interpreter-inspired models.}
    \label{tab:fusion-results}
    \centering
    \begin{center}
    \begin{small}
    \begin{sc}
    \resizebox{\textwidth}{!}{
\begin{tabular}{lllllllllllll}
\toprule
 {}    & \multicolumn{3}{l}{Baseline} & \multicolumn{3}{l}{Docstring}& \multicolumn{3}{l}{FiLM}    & \multicolumn{3}{l}{Cross-attention} \\
Model      &     Acc. & W. F1 & E. F1 &      Acc. & W. F1 & E. F1 &   Acc. & W. F1 & E. F1   &            Acc. & W. F1 & E. F1                \\
\midrule
IPA-GNN    &     68.3 &  64.8 &  53.8 &      71.4 &  70.1 &  62.2 &   71.6 &  70.3 &  62.9   &            72.0 &  70.3 &  62.6                \\
E. IPA-GNN &     68.7 &  64.9 &  53.3 &      71.6 &  70.9 &  63.5 &   70.9 &  68.8 &  59.8   &            73.8 &  72.3 &  64.7                \\
\bottomrule
\end{tabular}
    }
    \end{sc}
    \end{small}
    \end{center}
\end{table*}

\textbf{RQ2:} Across all architectures it is clear that external resource descriptions are essential for improved performance on the runtime error prediction task. On the IPA-GNN architectures, we see further improvements by considering architectures that incorporate the resource description directly into the execution step of the model, but these gains are inconsistent. Critically, using any resource description method is better than none at all.

To understand how the resource descriptions lead to better performance,
we compare in Figure~\ref{fig:visualization-instruction-pointers-p02753-s293274223} the instruction pointer values of two Exception IPA-GNN models on a single example (shown in Table~\ref{tab:visualization-example-p02753-s293274223}).
The model with the resource description recognizes that the \code{input()} calls will read input beyond the end of the stdin stream. In contrast, the model without the resource description has less reason to suspect an error would be raised by those calls. The descriptions of stdin in the Python Runtime Errors dataset also frequently reveal type information, expected ranges for numeric values, and formatting details about the inputs.
We visualize additional examples in Appendix~\ref{app:visualization}.

\subsection{Interpretability and Localization}
\label{sec:rq3}

We next investigate the behavior of the Exception IPA-GNN model,
    evaluating its ability to localize runtime errors without any localization supervision.
In unsupervised localization, the models predict the location of the error 
  despite being trained only with error presence and kind supervision.

\paragraph{Multiple Instance Learning Baselines}
The unsupervised localization task may be 
    viewed as multiple instance learning (MIL) \citep{MIL}.
    To illustrate, consider the subtask of predicting whether a particular line contains an error.
    In an $n$-line program, there are $n$ instances of this subtask.
    The available supervision only indicates if any one of these subtasks has an error, but does not specify which one. So, treating each subtask as an instance, the group of subtasks forms a bag of instances, and we view the setting as MIL.

We thus consider as baselines two variations on the Transformer architecture,
    each using multiple instance learning.
    The first is the ``Local MIL Transformer'', in which each statement in the program is encoded individually, as in the local node embeddings computation of Section~\ref{sec:per-node-embeddings}.
    The second is the ``Global MIL Transformer'', in which all tokens in the program may attend to all other tokens in the Transformer encoder.
    In both cases, the models make per-line predictions, which are aggregated to form
    an overall prediction as defined in Appendix~\ref{app:multiple-instance-learning}.

\paragraph{Localization Experiment}
We use the same experimental protocol as in Section~\ref{sec:rq1}, and train each of the MIL Transformer and Exception IPA-GNN models.
As before, the models are trained only to minimize cross-entropy loss on predicting error kind and presence, receiving no supervision as to the location of the errors.
We report the localization results in Table~\ref{tab:localization}.
Localization accuracy (``{\sc{Local.}}'' in the table) computes the percent of test examples with errors for which the model correctly predicts the line number of the error.

\begin{table}[h]
    \caption{Localization accuracy (\%) for the MIL Transformers and Exception IPA-GNN on the Python Runtime Errors balanced test split.}
    \label{tab:localization}
    \vspace{-12pt}
    \centering 
    \begin{center}
    \begin{small}
    \begin{sc}
    \begin{tabular}{lcc}
\toprule
Model                  & R.D.?    & Local. \\
\midrule
Local MIL Transformer  &                &    33.0  \\
Local MIL Transformer  & \Checkmark     &    48.9 \\
Global MIL Transformer &                &    48.2  \\
Global MIL Transformer & \Checkmark     &    48.8 \\
E. IPA-GNN             &                &    50.8  \\
E. IPA-GNN + Docstring & \Checkmark     &    64.7 \\
E. IPA-GNN + FiLM      & \Checkmark     &    64.5  \\
E. IPA-GNN + C.A.      & \Checkmark     &    \textbf{68.8}  \\
\bottomrule
\end{tabular}

    \end{sc}
    \end{small}
    \end{center}
\end{table}

\begin{figure*}[h!]
    \caption{
    Heatmap of instruction pointer values produced by \textsc{Baseline} and \textsc{Docstring} Exception IPA-GNNs for the example in Table~\ref{tab:visualization-example-p02753-s293274223}. The x-axis represents timesteps and the y-axis represents nodes, with the last two rows respectively representing $\nexit$ and $\nerror$.
    The \textsc{Baseline} instruction pointer value is diffuse, with most probability mass ending at $\nexit$. With the resource description, the instruction pointer value is sharp, with almost all probability mass jumping to $\nerror$ from node~2.
    }
    \label{fig:visualization-instruction-pointers}
    \label{fig:visualization-instruction-pointers-p02753-s293274223}
    \centering
    \begin{subfigure}[c]{0.48\columnwidth}
    \centering
    \includegraphics[width=\columnwidth]{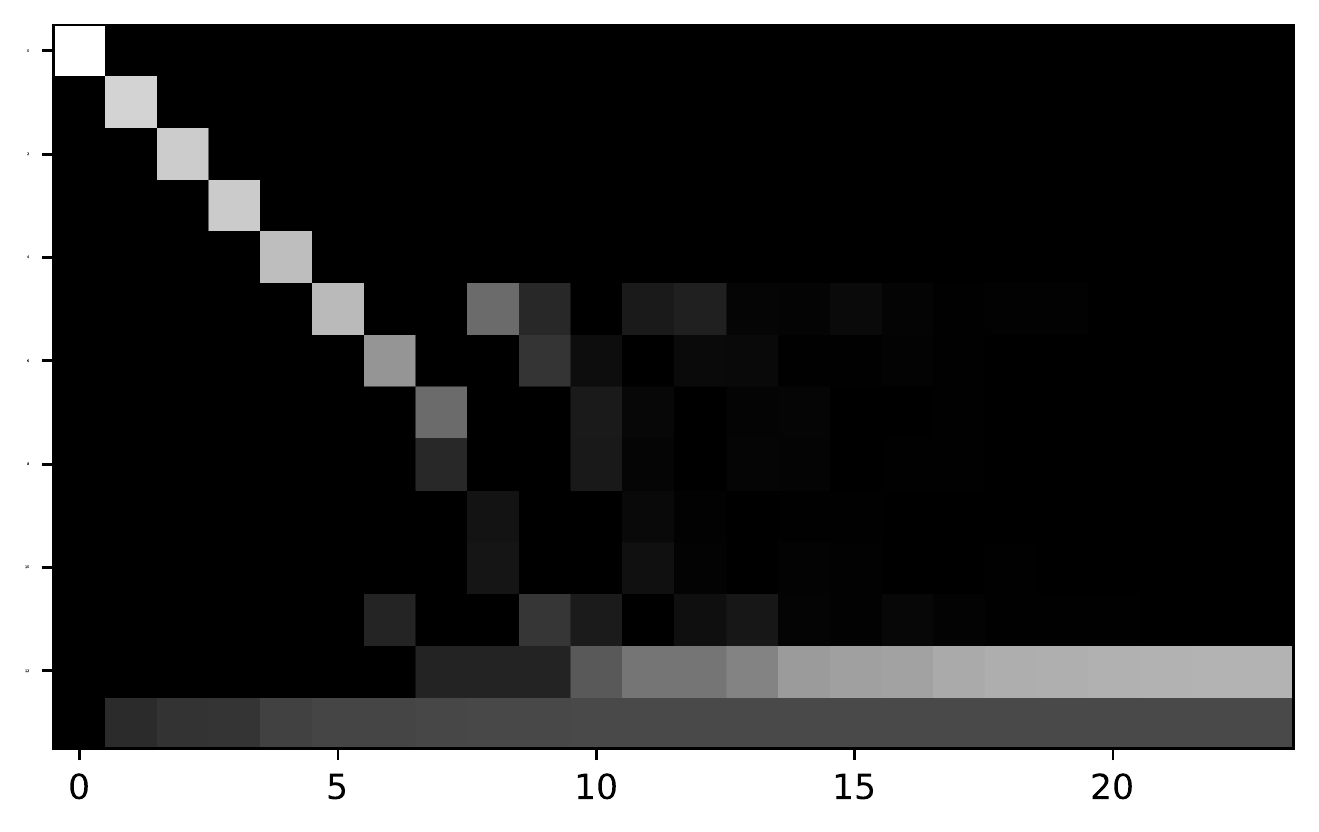}
    \caption*{\textsc{Baseline}}
    \end{subfigure}
    \begin{subfigure}[c]{0.48\columnwidth}
    \centering
    \includegraphics[width=\columnwidth]{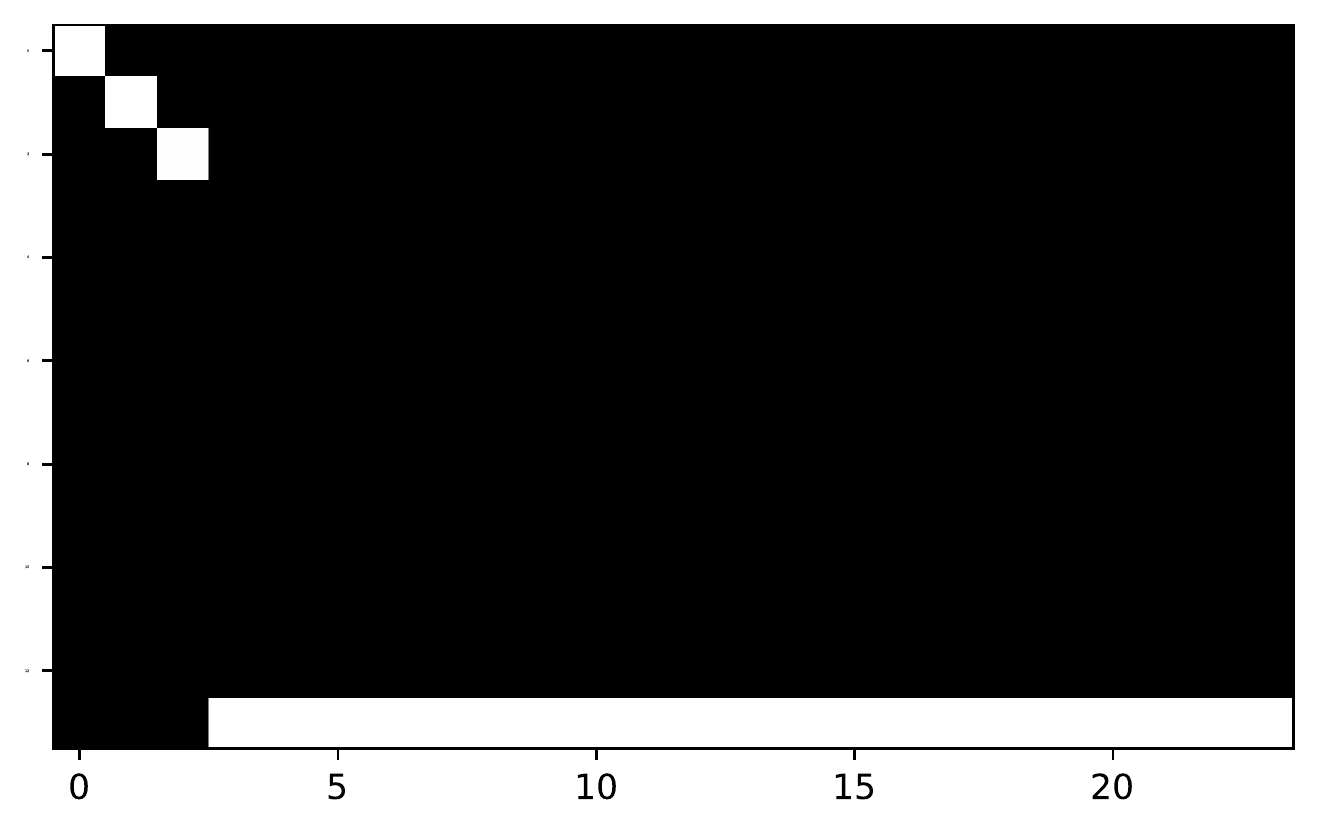}
    \caption*{\textsc{Resource Description}}
    \end{subfigure}
\end{figure*}

\textbf{RQ3:}
The Exception IPA-GNN's unsupervised localization capabilities far exceed that of baseline approaches.
In Figure~\ref{fig:visualization-instruction-pointers} we see the flow of instruction pointer 
mass during the execution of a sample program (Table~\ref{tab:visualization-example-p02753-s293274223}) by two Exception IPA-GNN models,
including the steps where the models raise probability mass to $\nerror$.
Tallying the contributions to $\nerror$ from each node yields the exception provenance values in the right half of Table~\ref{tab:visualization-example-p02753-s293274223}. This shows how the model's internal state resembles plausible program executions and allows for unsupervised localization.
As a beneficial side-effect of learning plausible executions, the Exception IPA-GNN can localize the exceptions it predicts.

\begin{table*}	
\caption{
    A sample program from the Python Runtime Errors validation split.
    Error node contributions are shown for the selected \textsc{Baseline} and \textsc{Docstring} Exception IPA-GNN variants.	
    The target error kind is \textsc{EOFError}, occuring on line 2 ($n = 2$). \textsc{Baseline} incorrectly predicts \textsc{No Error} with confidence 0.708. \textsc{R.D.} correctly predicts \textsc{EOFError} with confidence 0.988, localized at line 3 ($n = 3$). The input description shows the cause for error: there are more \texttt{input()} calls than the number of expected inputs (one).
}
\label{tab:visualization-example-p02753-s293274223}	

\begin{small}
\begin{subtable}{\textwidth}
\begin{tabularx}{\textwidth}{lX}
\textsc{StdIn Description} & \code{Input: Input is given from Standard Input in the following format      Constraints: Each character of S is A or B. |S| = 3}
\end{tabularx}
\end{subtable}

\begin{subtable}{\textwidth}
\begin{tabularx}{\textwidth}{cXrr}
\toprule
$n$ & \textsc{Source} & \textsc{Baseline} & \textsc{R.D.} \\
\midrule
\code{0}    & \code{a = str(input())}                                  & 16.9 &          0.4  \\
\code{1}    & \code{q = int(input())}                                  &  3.2 &          0.3  \\
\code{2}    & \code{s = [input().split() for i in range(q)]}           &  0.5 & \textbf{99.3} \\
\code{3,4}  & \code{for i in range(q):}                                &  6.4 &          0.0  \\
\code{5}    & \code{~~if int(s[i][0]) == 1 and len(a)\textgreater{}1:} &  0.1 &          0.0  \\
\code{6}    & \code{~~~~a = a[::-1]}                                   &  0.7 &          0.0  \\
\code{7}    & \code{~~elif int(s[i][0])== 2 and int(s[i][1])==1:}      &  0.1 &          0.0  \\
\code{8}    & \code{~~~~a=s[i][2]+a}                                   &  0.2 &          0.0  \\
            & \code{~~else:}                                           &      &               \\
\code{9}    & \code{~~~~a=a+s[i][2]}                                   &  0.0 &          0.0  \\
\code{10}   & \code{print(a)}                                          &  1.1 &          0.0  \\
\bottomrule
\end{tabularx}
\end{subtable}
\end{small}

\end{table*}

\setcounter{figure}{2}

\section{Discussion}

In this work, we introduce the new task of predicting runtime errors in competitive programming
problems and advance the capabilities of interpreter-inspired models.
We tackle the additional complexity of real-world code and demonstrate how natural language descriptions of external resources can be leveraged to reduce the ambiguity that arises in a static analysis setting.
We show that the resulting models outperform standard alternatives like Transformers and that the inductive bias built into these models allows for interesting interpretability in the context of unsupervised localization.

Though they perform best, current IPA-GNN models require taking many steps of execution, up to 174 on this dataset.
A future direction is to model multiple steps of program execution with a single model step, to reduce the number of model steps necessary for long programs.
Extending the interpreter-inspired models with additional interpreter features, or to support multi-file programs or programs with multiple user-defined functions is also an interesting avenue for future work.

Learning to understand programs remains a rich area of inquiry for machine learning research
because of its complexity and the many aspects of code. 
Learning to understand execution behavior is particularly challenging as programs grow in
complexity, and as they depend on more external resources whose contents are not present in the code.
Our work presents a challenging problem and advances interpreter-inspired models, both of
which we hope are ingredients towards making progress on these difficult and important problems.

\newpage

\bibliography{refs}

\begin{thebibliography}{37}
\providecommand{\natexlab}[1]{#1}
\providecommand{\url}[1]{\texttt{#1}}
\expandafter\ifx\csname urlstyle\endcsname\relax
  \providecommand{\doi}[1]{doi: #1}\else
  \providecommand{\doi}{doi: \begingroup \urlstyle{rm}\Url}\fi

\bibitem[Allamanis et~al.(2018{\natexlab{a}})Allamanis, Barr, Devanbu, and
  Sutton]{allamanis2018survey}
Miltiadis Allamanis, Earl~T Barr, Premkumar Devanbu, and Charles Sutton.
\newblock A survey of machine learning for big code and naturalness.
\newblock \emph{ACM Computing Surveys (CSUR)}, 51\penalty0 (4):\penalty0 81,
  2018{\natexlab{a}}.

\bibitem[Allamanis et~al.(2018{\natexlab{b}})Allamanis, Brockschmidt, and
  Khademi]{allamanis2018learning}
Miltiadis Allamanis, Marc Brockschmidt, and Mahmoud Khademi.
\newblock Learning to represent programs with graphs.
\newblock In \emph{International Conference on Learning Representations},
  2018{\natexlab{b}}.

\bibitem[Anonymous(2022)]{scratchpads}
Anonymous.
\newblock Show your work: Scratchpads for intermediate computation with
  language models.
\newblock In \emph{Submitted to The Tenth International Conference on Learning
  Representations}, 2022.
\newblock URL \url{https://openreview.net/forum?id=iedYJm92o0a}.
\newblock under review.

\bibitem[Austin et~al.(2021)Austin, Odena, Nye, Bosma, Michalewski, Dohan,
  Jiang, Cai, Terry, Le, and Sutton]{austin2021program}
Jacob Austin, Augustus Odena, Maxwell Nye, Maarten Bosma, Henryk Michalewski,
  David Dohan, Ellen Jiang, Carrie Cai, Michael Terry, Quoc Le, and Charles
  Sutton.
\newblock Program synthesis with large language models, 2021.

\bibitem[Ayewah et~al.(2008)Ayewah, Pugh, Hovemeyer, Morgenthaler, and
  Penix]{find-bugs}
Nathaniel Ayewah, William Pugh, David Hovemeyer, J.~David Morgenthaler, and
  John Penix.
\newblock Using static analysis to find bugs.
\newblock \emph{IEEE Software}, 25\penalty0 (5):\penalty0 22--29, 2008.
\newblock \doi{10.1109/MS.2008.130}.

\bibitem[Bieber et~al.(2020)Bieber, Sutton, Larochelle, and Tarlow]{ipagnn}
David Bieber, Charles Sutton, Hugo Larochelle, and Daniel Tarlow.
\newblock Learning to execute programs with instruction pointer attention graph
  neural networks.
\newblock In \emph{Advances in Neural Information Processing Systems}, 2020.

\bibitem[Bošnjak et~al.(2017)Bošnjak, Rocktäschel, Naradowsky, and
  Riedel]{differentiable-forth-interpreter}
Matko Bošnjak, Tim Rocktäschel, Jason Naradowsky, and Sebastian Riedel.
\newblock Programming with a differentiable forth interpreter, 2017.

\bibitem[Cadar et~al.(2008)Cadar, Dunbar, Engler, et~al.]{klee}
Cristian Cadar, Daniel Dunbar, Dawson~R Engler, et~al.
\newblock Klee: unassisted and automatic generation of high-coverage tests for
  complex systems programs.
\newblock In \emph{OSDI}, volume~8, pp.\  209--224, 2008.

\bibitem[Chen et~al.(2021{\natexlab{a}})Chen, Tworek, Jun, Yuan,
  de~Oliveira~Pinto, Kaplan, Edwards, Burda, Joseph, Brockman, Ray, Puri,
  Krueger, Petrov, Khlaaf, Sastry, Mishkin, Chan, Gray, Ryder, Pavlov, Power,
  Kaiser, Bavarian, Winter, Tillet, Such, Cummings, Plappert, Chantzis, Barnes,
  Herbert-Voss, Guss, Nichol, Paino, Tezak, Tang, Babuschkin, Balaji, Jain,
  Saunders, Hesse, Carr, Leike, Achiam, Misra, Morikawa, Radford, Knight,
  Brundage, Murati, Mayer, Welinder, McGrew, Amodei, McCandlish, Sutskever, and
  Zaremba]{codex}
Mark Chen, Jerry Tworek, Heewoo Jun, Qiming Yuan, Henrique~Ponde
  de~Oliveira~Pinto, Jared Kaplan, Harri Edwards, Yuri Burda, Nicholas Joseph,
  Greg Brockman, Alex Ray, Raul Puri, Gretchen Krueger, Michael Petrov, Heidy
  Khlaaf, Girish Sastry, Pamela Mishkin, Brooke Chan, Scott Gray, Nick Ryder,
  Mikhail Pavlov, Alethea Power, Lukasz Kaiser, Mohammad Bavarian, Clemens
  Winter, Philippe Tillet, Felipe~Petroski Such, Dave Cummings, Matthias
  Plappert, Fotios Chantzis, Elizabeth Barnes, Ariel Herbert-Voss,
  William~Hebgen Guss, Alex Nichol, Alex Paino, Nikolas Tezak, Jie Tang, Igor
  Babuschkin, Suchir Balaji, Shantanu Jain, William Saunders, Christopher
  Hesse, Andrew~N. Carr, Jan Leike, Josh Achiam, Vedant Misra, Evan Morikawa,
  Alec Radford, Matthew Knight, Miles Brundage, Mira Murati, Katie Mayer, Peter
  Welinder, Bob McGrew, Dario Amodei, Sam McCandlish, Ilya Sutskever, and
  Wojciech Zaremba.
\newblock Evaluating large language models trained on code, 2021{\natexlab{a}}.

\bibitem[Chen et~al.(2016)Chen, Xu, Zhang, and Guestrin]{deep-sublinear}
Tianqi Chen, Bing Xu, Chiyuan Zhang, and Carlos Guestrin.
\newblock Training deep nets with sublinear memory cost, 2016.

\bibitem[Chen et~al.(2021{\natexlab{b}})Chen, Hellendoorn, Maniatis, Lamblin,
  Manzagol, Tarlow, and Moitra]{plur}
Zimin Chen, Vincent~J Hellendoorn, Petros Maniatis, Pascal Lamblin,
  Pierre-Antoine Manzagol, Danny Tarlow, and Subhodeep Moitra.
\newblock Plur: A unifying, graph-based view of program learning,
  understanding, and repair.
\newblock 2021{\natexlab{b}}.

\bibitem[Dehghani et~al.(2019)Dehghani, Gouws, Vinyals, Uszkoreit, and Łukasz
  Kaiser]{universal-transformer}
Mostafa Dehghani, Stephan Gouws, Oriol Vinyals, Jakob Uszkoreit, and Łukasz
  Kaiser.
\newblock Universal transformers, 2019.

\bibitem[Dietterich et~al.(1997)Dietterich, Lathrop, and Lozano-Pérez]{MIL}
Thomas~G. Dietterich, Richard~H. Lathrop, and Tomás Lozano-Pérez.
\newblock Solving the multiple instance problem with axis-parallel rectangles.
\newblock \emph{Artificial Intelligence}, 89\penalty0 (1):\penalty0 31--71,
  1997.
\newblock ISSN 0004-3702.
\newblock \doi{https://doi.org/10.1016/S0004-3702(96)00034-3}.
\newblock URL
  \url{https://www.sciencedirect.com/science/article/pii/S0004370296000343}.

\bibitem[Gaunt et~al.(2017)Gaunt, Brockschmidt, Kushman, and
  Tarlow]{gaunt2017differentiable}
Alexander~L. Gaunt, Marc Brockschmidt, Nate Kushman, and Daniel Tarlow.
\newblock Differentiable programs with neural libraries, 2017.

\bibitem[Godefroid et~al.(2005)Godefroid, Klarlund, and Sen]{godefroid2005dart}
Patrice Godefroid, Nils Klarlund, and Koushik Sen.
\newblock Dart: Directed automated random testing.
\newblock In \emph{Proceedings of the 2005 ACM SIGPLAN conference on
  Programming language design and implementation}, pp.\  213--223, 2005.

\bibitem[Graves et~al.(2014)Graves, Wayne, and
  Danihelka]{neural-turing-machines}
Alex Graves, Greg Wayne, and Ivo Danihelka.
\newblock Neural turing machines, 2014.

\bibitem[Graves et~al.(2016)Graves, Wayne, Reynolds, Harley, Danihelka,
  Grabska-Barwińska, Colmenarejo, Grefenstette, Ramalho, Agapiou, Badia,
  Hermann, Zwols, Ostrovski, Cain, King, Summerfield, Blunsom, Kavukcuoglu, and
  Hassabis]{differentiable-neural-computers}
Alex Graves, Greg Wayne, Malcolm Reynolds, Tim Harley, Ivo Danihelka, Agnieszka
  Grabska-Barwińska, Sergio~Gómez Colmenarejo, Edward Grefenstette, Tiago
  Ramalho, John Agapiou, Adrià~Puigdomènech Badia, Karl~Moritz Hermann, Yori
  Zwols, Georg Ostrovski, Adam Cain, Helen King, Christopher Summerfield, Phil
  Blunsom, Koray Kavukcuoglu, and Demis Hassabis.
\newblock Hybrid computing using a neural network with dynamic external memory.
\newblock \emph{Nature}, 538\penalty0 (7626):\penalty0 471--476, October 2016.
\newblock ISSN 00280836.
\newblock URL \url{http://dx.doi.org/10.1038/nature20101}.

\bibitem[Griewank \& Walther(2000)Griewank and Walther]{gradient-checkpointing}
Andreas Griewank and Andrea Walther.
\newblock Algorithm 799: Revolve: An implementation of checkpointing for the
  reverse or adjoint mode of computational differentiation.
\newblock \emph{ACM Trans. Math. Softw.}, 26\penalty0 (1):\penalty0 19–45,
  mar 2000.
\newblock ISSN 0098-3500.
\newblock \doi{10.1145/347837.347846}.
\newblock URL \url{https://doi.org/10.1145/347837.347846}.

\bibitem[Hellendoorn et~al.(2020)Hellendoorn, Sutton, Singh, and
  Maniatis]{great-paper}
Vincent~J Hellendoorn, Charles Sutton, Rishabh Singh, and Petros Maniatis.
\newblock Global relational models of source code.
\newblock In \emph{International Conference on Learning Representations}, 2020.

\bibitem[Hochreiter \& Schmidhuber(1997)Hochreiter and Schmidhuber]{lstm}
Sepp Hochreiter and J\"{u}rgen Schmidhuber.
\newblock Long short-term memory.
\newblock \emph{Neural Comput.}, 9\penalty0 (8):\penalty0 1735–1780, November
  1997.
\newblock ISSN 0899-7667.
\newblock \doi{10.1162/neco.1997.9.8.1735}.
\newblock URL \url{https://doi.org/10.1162/neco.1997.9.8.1735}.

\bibitem[Karampatsis \& Sutton(2020)Karampatsis and Sutton]{sstubs}
Rafael-Michael Karampatsis and Charles Sutton.
\newblock How often do single-statement bugs occur?
\newblock \emph{Proceedings of the 17th International Conference on Mining
  Software Repositories}, Jun 2020.
\newblock \doi{10.1145/3379597.3387491}.
\newblock URL \url{http://dx.doi.org/10.1145/3379597.3387491}.

\bibitem[Lee et~al.(2019)Lee, Lee, Kim, Kosiorek, Choi, and
  Teh]{cross-attention}
Juho Lee, Yoonho Lee, Jungtaek Kim, Adam~R. Kosiorek, Seungjin Choi, and
  Yee~Whye Teh.
\newblock Set transformer: A framework for attention-based
  permutation-invariant neural networks, 2019.

\bibitem[Li et~al.(2017)Li, Tarlow, Brockschmidt, and Zemel]{ggnn}
Yujia Li, Daniel Tarlow, Marc Brockschmidt, and Richard Zemel.
\newblock Gated graph sequence neural networks, 2017.

\bibitem[Livshits \& Lam(2005)Livshits and Lam]{livshits2005finding}
V~Benjamin Livshits and Monica~S Lam.
\newblock Finding security vulnerabilities in java applications with static
  analysis.
\newblock In \emph{USENIX security symposium}, volume~14, pp.\  18--18, 2005.

\bibitem[Nielson \& Nielson(1999)Nielson and Nielson]{icfg}
Flemming Nielson and Hanne~Riis Nielson.
\newblock Interprocedural control flow analysis.
\newblock In \emph{ESOP}, 1999.

\bibitem[Perez et~al.(2017)Perez, Strub, de~Vries, Dumoulin, and
  Courville]{film}
Ethan Perez, Florian Strub, Harm de~Vries, Vincent Dumoulin, and Aaron
  Courville.
\newblock Film: Visual reasoning with a general conditioning layer, 2017.

\bibitem[Puri et~al.(2021)Puri, Kung, Janssen, Zhang, Domeniconi, Zolotov,
  Dolby, Chen, Choudhury, Decker, Thost, Buratti, Pujar, Ramji, Finkler,
  Malaika, and Reiss]{project-codenet}
Ruchir Puri, David~S. Kung, Geert Janssen, Wei Zhang, Giacomo Domeniconi,
  Vladimir Zolotov, Julian Dolby, Jie Chen, Mihir Choudhury, Lindsey Decker,
  Veronika Thost, Luca Buratti, Saurabh Pujar, Shyam Ramji, Ulrich Finkler,
  Susan Malaika, and Frederick Reiss.
\newblock Codenet: A large-scale ai for code dataset for learning a diversity
  of coding tasks, 2021.

\bibitem[Reed \& de~Freitas(2016)Reed and
  de~Freitas]{neural-programmer-interpreters}
Scott Reed and Nando de~Freitas.
\newblock Neural programmer-interpreters, 2016.

\bibitem[Sen et~al.(2005)Sen, Marinov, and Agha]{sen2005cute}
Koushik Sen, Darko Marinov, and Gul Agha.
\newblock Cute: A concolic unit testing engine for c.
\newblock \emph{ACM SIGSOFT Software Engineering Notes}, 30\penalty0
  (5):\penalty0 263--272, 2005.

\bibitem[Sennrich et~al.(2016)Sennrich, Haddow, and Birch]{bytepairencoding}
Rico Sennrich, Barry Haddow, and Alexandra Birch.
\newblock Neural machine translation of rare words with subword units, 2016.

\bibitem[Vasic et~al.(2019)Vasic, Kanade, Maniatis, Bieber, and
  Singh]{vasic-localize-and-repair}
Marko Vasic, Aditya Kanade, Petros Maniatis, David Bieber, and Rishabh Singh.
\newblock Neural program repair by jointly learning to localize and repair.
\newblock In \emph{International Conference on Learning Representations}, 2019.

\bibitem[{Vaswani} et~al.(2017){Vaswani}, {Shazeer}, {Parmar}, {Uszkoreit},
  {Jones}, {Gomez}, {Kaiser}, and {Polosukhin}]{transformer}
Ashish {Vaswani}, Noam {Shazeer}, Niki {Parmar}, Jakob {Uszkoreit}, Llion
  {Jones}, Aidan~N. {Gomez}, Lukasz {Kaiser}, and Illia {Polosukhin}.
\newblock {Attention Is All You Need}.
\newblock \emph{arXiv e-prints}, art. arXiv:1706.03762, Jun 2017.

\bibitem[Wang et~al.(2021)Wang, Liu, Lin, Li, Klein, Mao, and
  Bissyandé]{beep-localization}
Shangwen Wang, Kui Liu, Bo~Lin, Li~Li, Jacques Klein, Xiaoguang Mao, and
  Tegawendé~F. Bissyandé.
\newblock Beep: Fine-grained fix localization by learning to predict buggy code
  elements, 2021.

\bibitem[Wang et~al.(2018)Wang, Li, and Metze]{maxAndNoisyOrMIL}
Yun Wang, Juncheng Li, and Florian Metze.
\newblock Comparing the max and noisy-or pooling functions in multiple instance
  learning for weakly supervised sequence learning tasks, 2018.

\bibitem[Xie \& Aiken(2006)Xie and Aiken]{xie2006static}
Yichen Xie and Alex Aiken.
\newblock Static detection of security vulnerabilities in scripting languages.
\newblock In \emph{USENIX Security Symposium}, volume~15, pp.\  179--192, 2006.

\bibitem[Zaremba \& Sutskever(2014)Zaremba and Sutskever]{learning-to-execute}
Wojciech Zaremba and Ilya Sutskever.
\newblock Learning to execute, 2014.

\bibitem[Łukasz Kaiser \& Sutskever(2016)Łukasz Kaiser and
  Sutskever]{neural-gpu}
Łukasz Kaiser and Ilya Sutskever.
\newblock Neural gpus learn algorithms, 2016.

\end{thebibliography}
\bibliographystyle{iclr2022_conference}

\newpage

\appendix

\section{Python Runtime Error Dataset Details}

\label{app:dataset-details}
We describe in detail the construction of the Python Runtime Error dataset
    from the submissions in \codenet \citep{project-codenet}.
The \codenet dataset contains
    over 14 million submissions
    to 4,053 distinct competitive programming problems,
    with the submissions spanning more than 50 programming languages.
We partition the problems into train, valid, and test splits at an 80:10:10 ratio.
By making all submissions to the same problem part of the same split
    we mitigate concerns about potential data leakage from similar submissions to the same problem.
We restrict our consideration to Python submissions,
    which account for 3,286,314 of the overall \codenet submissions,
    with 3,119 of the problems receiving at least one submission in Python.
In preparing the dataset
    we execute approximately 3 million problems in a sandboxed environment to collect their runtime error information,
    we perform two stages of filtering on the dataset,
    syntactic and complexity filtering,
    and we construct a textual representation of the input space for each problem from the problem description.

\subsection{Syntactic Filtering}

In this first phase of filtering, we remove submissions in Python 2 as well as those which fail to parse and run from our dataset.
We remove 76,888 programs because they are in Python 2,
    59,813 programs because they contain syntax errors,
    2,011 programs that result in runtime errors during parsing,
    and 6 additional programs for which the python-graphs library fails to construct a control flow graph.
A program may result in a runtime error during parsing if it contains return, break, continue keywords outside of an appropriate frame.

\subsection{Program Execution}

We attempt to run each submission in a sandboxed environment
    using the sample input provided in the \codenet dataset.
The environment is a custom harness running on a Google Cloud Platform (GCP) virtual environment.
This allows us to collect standard out and standard error,
    to monitor for timeouts,
    and to catch and serialize any Python exceptions raised during execution.
We restrict execution of each program to 1 second, marking any program exceeding this time as a timeout error.
If the program encounters a Python exception, we use the name of that exception
    as the target class for the program.
If an error type occurs only once in the dataset, we consider the target class to be Other.
Programs not exhibiting an error or timeout are given target class ``no error''.

In addition to collecting the target class,
    we record for each runtime error the line number at which the error occurs.
We use these line numbers as the ground truth for the unsupervised error localization task considered in Section~\ref{sec:rq3}.

\subsection{Parsing Input Space Information from Problem Statements}

For each problem, we parse the problem statement
    to extract the \emph{input description} and \emph{input constraints}, if they exist.
Problem statements are written either in English or Japanese,
    and so we write our parser to support both languages.
When one or more of these two sections are present in the problem statement,
    we construct an \emph{input space description} containing the contents of the present sections.
For the experiments that use input space information as a docstring,
    we prepend to each submission our the input space description for its corresponding problem.
Similarly the input space descriptions are used in the experiments that process input space information with either cross-attention or FiLM.

\subsection{Vocabulary Construction and Complexity Filtering}

All experiments use the same vocabulary and tokenization procedure.
For this, we select the standard Byte-Pair Encoding (BPE) tokenization procedure \citep{bytepairencoding}.
We construct the vocabulary using 1,000,000 submissions selected from the training split,
    along with the input space descriptions constructed for all problems in the train split.
We use a vocabulary size of 30,000.

We then apply size-based filtering, further restricting the set of programs considered.
First, the program length after tokenization is not to exceed 512 tokens,
    the number of nodes and edges in the control flow graph are each not to exceed 128,
    and the step limit $T(x)$ for a program computed in Appendix~\ref{app:step-limit} is not to exceed 174.
We select these numbers to trim the long tail of exceptionally long programs, and this filtering reduces the total number of acceptable programs by less than 1\%.
To achieve consistent datasets comparable across all experiments,
  we use the longest form of each program
  (the program augmented with its input space information as a docstring),
  when computing the program sizes for size-based submission filtering.

We further impose the restriction that no user-defined functions (UDFs) are called in a submission;
  this further reduces the number of submissions by 682,220.
A user-defined function is a function defined in the submission source code,
  as opposed to being a built-in or imported from a third party module.
Extending the IPA-GNN models to submissions with UDFs called \emph{at most once}
  is trivially achieved by replacing the program's control flow graph with
  its interprocedural control flow graph (ICFG) \citep{icfg}.
We leave the investigation of modelling user-defined functions to further work. 

\subsection{Final Dataset Details}

\begin{figure}
\RawFloats
\centering
\begin{minipage}{.48\textwidth}
  \centering
  \includegraphics[width=\linewidth]{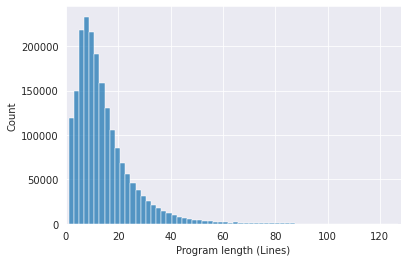}
  \captionof{figure}{A histogram showing the distribution of program lengths, measured in lines, represented in the Python Runtime Errors train split.}
  \label{fig:program-lengths}
\end{minipage}%
\hfill
\begin{minipage}{.47\textwidth}
  \centering
  \includegraphics[width=\linewidth]{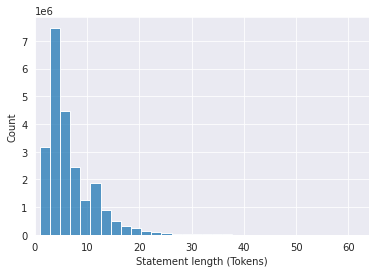}
  \captionof{figure}{The distribution of statement lengths, measured in tokens, in the Python Runtime Errors train split.}
  \label{fig:statement-lengths}
\end{minipage}
\end{figure}

After applying syntactic filtering (only keeping Python 3 programs that parse)
    and complexity filtering (eliminating long programs and programs that call user-defined functions), we are left with a dataset of 2,441,130 examples.
The division of these examples by split and by target class is given in Table~\ref{tab:error-frequencies}. Figure~\ref{fig:program-lengths} shows the distribution of program lengths in lines represented in the completed dataset, with an average program length of 14.2 lines. The average statement length is 6.7 tokens, with full distribution shown in Figure~\ref{fig:statement-lengths}.

\section{Input Modulation}
\label{app:modulation-function}
We consider \emph{cross-attention} \citep{cross-attention} and \emph{Feature-wise Linear Modulation} (FiLM) \citep{film} as the $\Modulate$ function. After embedding the node and the resource description we use cross-attention as follows to modulate the input.
\begin{align}
\MultiHead(\Embed(x_n), d(x), h_{t-1, n}) &=  \Concat(\Concat(head_1, ..., head_h) W^O,  \Embed(x_n))\\
\text{where\ }head_i &= \Softmax\left(\frac{QK^{'}}{\sqrt{d_k}}\right)V\\
Q &=  W^Q_i \Concat(Embed(x_n), h_{t-1, n})\\
K &=  W^K_i d(x)\\
V &=  W^V_i d(x)
\end{align}
Here,
$W^O \in R^{hd_{v} \times d_{model}}$,
$W^Q_i \in R^{d_k \times (d_{model} + d_{\Embed(x_n)})}$,
$W^K_i \in R^{ d_k  \times d_{d(x)}}$,
and $W^V_i \in R^{d_v \times d_{d(x)}}$ are learnable parameters.
Similarly, for FiLM we modulate the input with the resource description as follows:
\begin{align}
    \film(\Embed(x_n), d(x), h_{t-1, n}) &= \Concat(\beta \cdot d(x) + \gamma, \Embed(x_n)) \\
    \text{where\ }\beta &= \sigma(W_{\beta} \Concat(x_n,h_{t-1, n})  + b_{\beta}),\\
    \gamma &= \sigma(W_{\gamma} \Concat(x_n,h_{t-1, n}) + b_{\gamma}),
\end{align}
where $W_{\gamma} \in R^{d_{d(x)} \times (d_{model} + d_{\Embed(x_n)})}$, and $W_{\gamma} \in R^{d_{d(x)} \times (d_{model} + d_{\Embed(x_n)})}$ are learnable parameters.

\section{Hyperparameter Selection}
\label{app:baseline-hparams}

We select hyperparameters by performing a random search independently for each model architecture.
The hyperparameters considered by the search are listed in Table~\ref{tab:hparams}.
All architectures use a Transformer encoder, and the Transformer sizes considered in the search are listed in Table~\ref{tab:hparams} and defined further in Table~\ref{tab:transformer-definitions}.

\begin{table}[h]
    \centering
    \begin{center}
    \begin{small}
    \begin{sc}
    \begin{tabular}{rlll}
    \toprule
        Hyperparameter & T-128 & T-256 & T-512 \\
        \midrule
        Embedding dimension &  128  &  256  & 512 \\
        Number of heads     &    4  &    4  &   8 \\
        Number of layers    &    2  &    2  &   6 \\
        qkv dimension       &  128  &  256  & 512 \\
        MLP dimension       &  512  & 1024  & 2048 \\
        \bottomrule
    \end{tabular}
    \end{sc}
    \end{small}
    \end{center}
    \caption{Hyperparameter settings for each of the three Transformer sizes.}
    \label{tab:transformer-definitions}
\end{table}

\label{app:step-limit}
For the IPA-GNN and Exception IPA-GNN, the function $T(x)$ represents the number of execution steps modeled for program $x$. We reuse the definition of $T(x)$ from \citet{ipagnn} as closely as possible, only modifying it to accept arbitrary Python programs, rather than being restricted to the subset of Python features considered in the dataset of the earlier work.

\begin{table}[b]
\centering
    \begin{center}
    \begin{small}
    \begin{sc}
\resizebox{\textwidth}{!}{
\begin{tabular}{rll}
    \toprule
        Hyperparameter & Value(s) considered & Architecture(s) \\
        \midrule
        Optimizer         & \code{\{SGD\}}  & All\\
        Batch size        & \code{\{32\}}   & All\\
        Learning rate     & \code{\{0.01, 0.03, 0.1, 0.3\}}   & LSTM, Transformers, IPA-GNNs\\
        Learning rate     & \code{\{0.001, 0.003, 0.01, 0.03\}}   & GGNN\\
        Gradient clipping & \code{\{0, 0.5, 1, 2\}}   & All\\
        Hidden size & \code{\{64, 128, 256\}}   & All\\
        RNN layers & \code{\{2\}}   & LSTM, IPA-GNNs\\
        GNN layers & \code{\{8, 16, 24\}}   & GGNN\\
        Span encoder pooling & \code{\{first, mean, max, sum\}}   & All\\
        Cross-attention number of heads & \code{\{1, 2\}}   & IPA-GNNs with Cross-attention\\
        MIL pooling & \code{\{max, mean, logsumexp\}}   & MIL Transformers\\
        Transformer dropout rate & \code{\{0, 0.1\}}   & All\\
        Transformer attention dropout rate & \code{\{0, 0.1\}}   & All\\
        Transformer size & \code{\{T-128, T-256, T-512\}}   & All\\
        \bottomrule
    \end{tabular}
    }
    \end{sc}
    \end{small}
    \end{center}
    \caption{Hyperparameters considered for random search during model selection.}
    \label{tab:hparams}
\end{table}

\section{Localization by Modeling Exception Handling}
\label{app:localization}

For programs that lack try/except frames, we compute the localization predictions of the Exception IPA-GNN model by summing,
    separately for each node, the contributions from that node to the exception node across all time steps.
This gives an estimate of exception provenance as
\begin{equation}
    p(\text{error at statement }n) = \sum_{t} {p_{t,n} \cdot b_{t,n,\nerror}}.
\end{equation}
For programs with a try/except frame, however, we must trace the exception back to the statement that originally raised it. To do this, we keep track of the exception provenance at each node at each time step; when an exception raises, it becomes the exception provenance at the statement that it raises to, and when a statement with non-zero exception provenance executes without raising, it propagates its exception provenance to the next node unchanged.

Define $v_{t,n,n'}$ as the amount of ``exception probability mass'' at time step $t$
  at node $n'$ attributable to an exception starting at node $n$.
Then we write
\begin{equation}
    v_{t,n,n'} = \sum_{k \in \NeighborsIn(n')} v_{t-1,n,k} \cdot b_{t, k, n'} \cdot p_{t, k}
    + (1 - \sum v_{t-1,:,n}) \cdot b_{t, n, n'} \cdot p_{t, n} \cdot \mathds{1}\{n'=r(n)\}.
\end{equation}
The first term propagates exception provenance across normal non-raising execution,
while the second term introduces exception provenance when an exception is raised.
We then write precisely
\begin{equation}
    p(\text{error at statement }n) = v_{T(x), n, \nerror},
\end{equation}
allowing the Exception IPA-GNN to make localization predictions for any program in the dataset.

\section{Metric Variances}
\label{app:variance}

Under the experimental conditions of Section~\ref{sec:rq1},
we perform three additional training runs to calculate the variance for each metric for each baseline model, and for the Exception IPA-GNN model using the docstring strategy for processing resource descriptions.
For these new training runs, we use the hyperparameters obtained from model selection.
We vary the random seed between runs (0, 1, 2), thereby changing the initialization and dropout behavior of each model across runs. We report the results in Table~\ref{tab:variance-results}; $\pm$ values are one standard deviation.

\begin{table}[h]
    \caption{Mean and standard deviation for each metric is calculated from three training runs per model, using the hyperparameters selected via model selection.}
    \label{tab:variance-results}
    \vspace{-12pt}
    \centering 
    \begin{center}
    \begin{small}
    \begin{sc}
    \begin{tabular}{rcccc}
\toprule
                               Method & R.D.?    &  Acc. &  W. F1 &  E. F1  \\
\midrule
                                 GGNN &                & 61.98 $\pm$  1.24 &  56.62  $\pm$ 2.96 &  41.24  $\pm$ 5.51   \\
                          Transformer &                & 63.82 $\pm$  0.62 &  59.86  $\pm$ 0.52 &  46.75  $\pm$ 0.93   \\
                                 LSTM &                & 66.43 $\pm$  0.60 &  62.33  $\pm$ 1.12 &  50.10  $\pm$ 1.94   \\
                    Exception IPA-GNN &\CheckmarkBold  & 71.44 $\pm$  0.15 &  70.78  $\pm$ 0.07 &  63.54  $\pm$ 0.03   \\
\bottomrule
\end{tabular}
    \end{sc}
    \end{small}
    \end{center}
\end{table}

\section{Multiple Instance Learning}
\label{app:multiple-instance-learning}

The Local Transformer and Global Transformer models
    each compute per-statement node embeddings $\Embed(x_n)$ given by Equation~\ref{equ:embed}.
In the multiple instance learning setting, these are transformed into unnormalized per-statement class predictions
\begin{equation}
    \phi(\text{class = }k, \text{lineno = }l) =
    \Dense\left(\Embed(x_n)\right).
\end{equation}
We consider three strategies for aggregating these per-statement predictions into an overall prediction for the task.
Under the \emph{logsumexp} strategy, we treat $\phi$ as logits and write
\begin{align}
    \log p(\text{class = }k) &\propto \log\left(\sum_{l}\exp{\phi(k, l)}\right), \\
    \log p(\text{lineno = }l) &\propto \log\left(\sum_{k \in K}\exp{\phi(k, l)}\right)
\end{align}
where K is the set of error classes.

The \emph{max} and \emph{mean} strategies meanwhile follow \citet{maxAndNoisyOrMIL} in asserting
\begin{align}
    p(\text{class = }k \mid \text{lineno = }l) &= \Softmax\left(\phi(k, l)\right),
\end{align}
compute the location probabilities as
\begin{align}
    p(\text{lineno = }l) &\propto \sum_{k \in K} p(\text{class = }k \mid \text{lineno = }l),
\end{align}
and compute the outputs as
\begin{align}
    \log p(\text{class = }k) &\propto \log \max_{l}{p(\text{class = }k \mid \text{lineno = }l)}\text{, and} \\
    \log p(\text{class = }k) &\propto \log \frac{1}{L}\sum_{l}{p(\text{class = }k \mid \text{lineno = }l)}
\end{align}
respectively, where $L$ denotes the number of statements in $x$.
As with all methods considered,
the MIL models are trained to minimize the cross-entropy loss in target class prediction,
but these methods still allow reading off predictions of $p(\text{lineno})$.

\section{Example visualizations}
\label{app:visualization}

Additional randomly sampled examples from the Python Runtime Error dataset validation split are shown here. As in Figure~\ref{fig:visualization-instruction-pointers}, prediction visualizations for these examples are shown for the selected \textsc{Baseline} and \textsc{Docstring} Exception IPA-GNN model variants.

In instruction pointer value heatmaps, the x-axis represents timesteps and the y-axis represents nodes, with the last two rows respectively representing the exit node $\nexit$ and the exception node $\nerror$.
Note that for loop statements are associated with two spans in the statement-level control flow graph, one for the construction of the iterator, and a second for assignment to the loop variable. Hence we list two indexes for each for loop statement in these figures, and report the total error contribution for the line.

\begin{figure*}[h]
\caption{
    The target error kind is \textsc{IndexError}, occuring on line 5 ($n = 5$). \textsc{Baseline} incorrectly predicts \textsc{No error} with confidence 0.808. \textsc{Docstring} correctly predicts \textsc{IndexError} with confidence 0.693, but localizes to line 3 ($n = 2$).
    Both \textsc{Baseline} and \textsc{Docstring} instruction pointer values start out sharp and become diffuse when reaching the for-loop. The \textsc{Baseline} instruction pointer value ends with most probability mass at $\nexit$. The \textsc{Docstring} instruction pointer value has a small amount of probability mass reaching $\nexit$, with most probability mass ending at $\nerror$.
}
\label{fig:visualization-example-p02607-s841000725}
\begin{small}
\begin{subtable}{\textwidth}
\begin{tabularx}{\textwidth}{lX}
\multirow{2}{*}{\textsc{StdIn Description}}
& \code{Input: Input is given from Standard Input in the following format: N a\_1 a\_2 ... a\_N} \\
& \code{Constraints: All values in input are integers. 1 <= N , a\_i <= 100}
\end{tabularx}
\end{subtable}
\begin{subtable}{\textwidth}
\begin{tabularx}{\textwidth}{cXrr}
\toprule
$n$ & \textsc{Source} & \textsc{Baseline} Error contrib. & \textsc{R.D.} Error contrib. \\
\midrule
\code{0}   & \code{N = int(input())}                    &  2.9 &          0.2 \\
\code{1}   & \code{A = list(map(int, input().split()))} &  0.8 &          0.0 \\
\code{2}   & \code{res = 0}                             &  3.0 & \textbf{63.3} \\
\code{3,4} & \code{for i in range(1, len(A)+1, 2):}     &  9.8 &          6.3 \\
\code{5}   & \code{~~res += A[i] \% 2}                  &  0.3 &          0.1 \\
\code{6}   & \code{print(res)}                          &  0.2 &          2.2 \\
\bottomrule
\end{tabularx}
\end{subtable}
\end{small}

\vspace{-10px}
\begin{subfigure}[c]{\textwidth}
\begin{subfigure}[c]{0.48\textwidth}
\centering
\includegraphics[width=\textwidth]{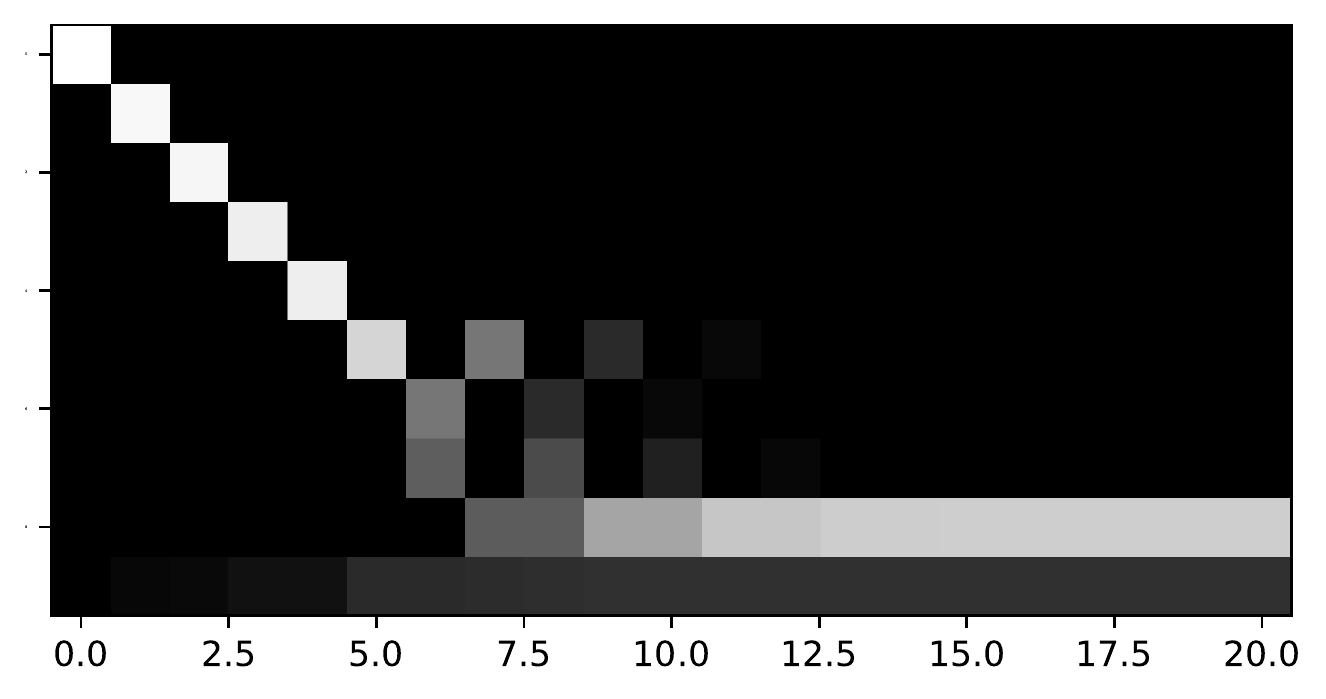}
\caption*{\textsc{Baseline}}
\end{subfigure}
\begin{subfigure}[c]{0.48\textwidth}
\centering
\includegraphics[width=\textwidth]{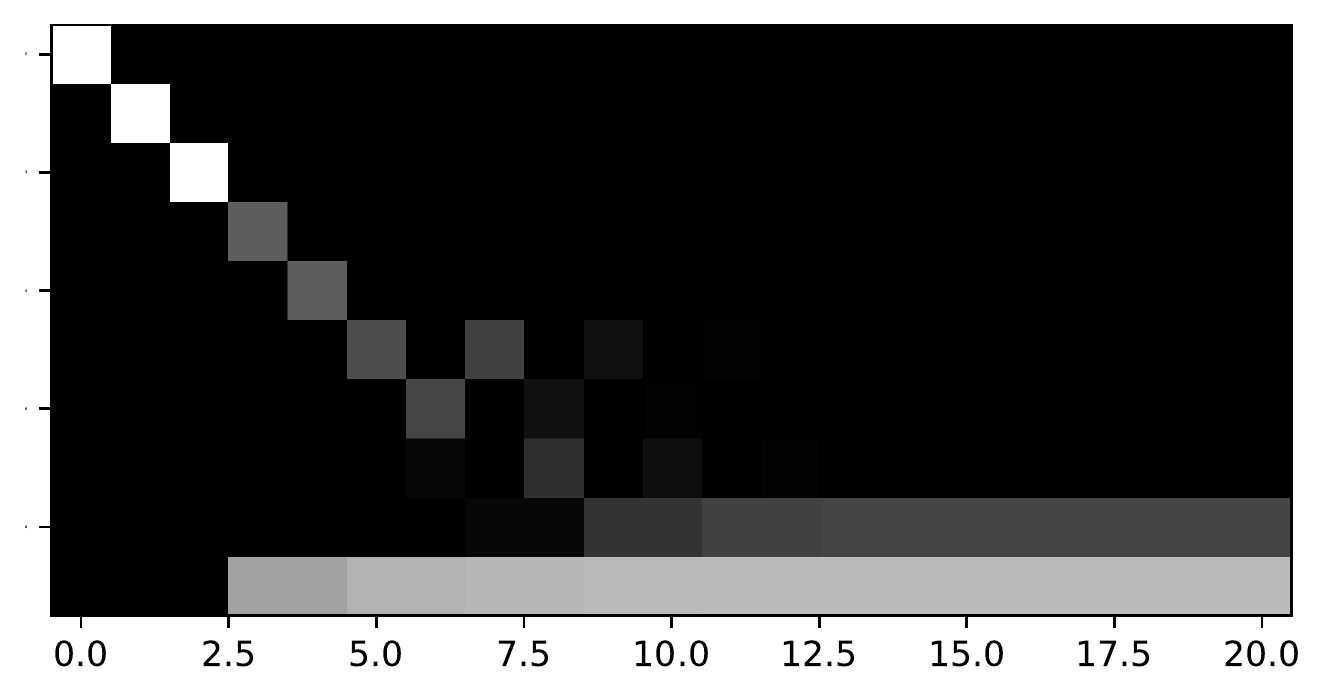}
\caption*{\textsc{Resource Description}}
\end{subfigure}
\end{subfigure}
\end{figure*}
\setcounter{figure}{5}

\begin{figure*}[h]
\caption{
    The target error kind is \textsc{ValueError}, occuring on line 1 ($n = 0$). \textsc{Baseline} incorrectly predicts \textsc{IndexError} with confidence 0.319 on line 1 ($n = 0$). \textsc{Docstring} correctly predicts \textsc{ValueError} with confidence 0.880 on line 2 ($n = 1$), corresponding to \code{A[n]}.
    Both \textsc{Baseline} and \textsc{Docstring} instruction pointer values start out sharp and quickly shift most of the probability mass to the exception node.
}
\label{tab:visualization-example-p02784-s135671180}

\begin{small}
\begin{subtable}{\textwidth}
\begin{tabularx}{\textwidth}{lX}
\multirow{3}{*}{\textsc{StdIn Description}}
& \code{Input: Input is given from Standard Input in the following format: H N A\_1 A\_2 ... A\_N} \\
& \code{Constraints: 1 <= H <= 10\^{}9 1 <= N <= 10\^{}5 1 <= A\_i <= 10\^{}4} \\
& \code{All values in input are integers.}
\end{tabularx}
\end{subtable}

\begin{subtable}{\textwidth}
\begin{tabularx}{\textwidth}{cXrr}
\toprule
$n$ & \textsc{Source} & \textsc{Baseline} Error contrib. & \textsc{R.D.} Error contrib. \\
\midrule
\code{0}   & \code{H,N,A = list(map(int, input().split()))} &          9.7  &          3.4  \\
\code{1,2} & \code{for i in A[N]:}                          & \textbf{43.7} & \textbf{83.0} \\
\code{3}   & \code{~~if H \textless{}= 0:}                  &          2.9  &          2.8  \\
           & \code{~~~~break}                               &               &               \\
           & \code{~~else:}                                 &               &               \\
\code{4}   & \code{~~~~H -= A[i]}                           &          6.0  &          0.0  \\
\code{5}   & \code{if set(A):}                              &          0.2  &          0.1  \\
\code{6}   & \code{~~print("Yes")}                          &          9.3  &          0.7  \\
           & \code{else:}                                   &               &               \\
\code{7}   & \code{~~print("No")}                           &          3.3  &          0.2  \\
\bottomrule
\end{tabularx}
\end{subtable}
\end{small}

\begin{subfigure}[c]{\textwidth}
\begin{subfigure}[c]{0.48\textwidth}
\centering
\includegraphics[width=\textwidth]{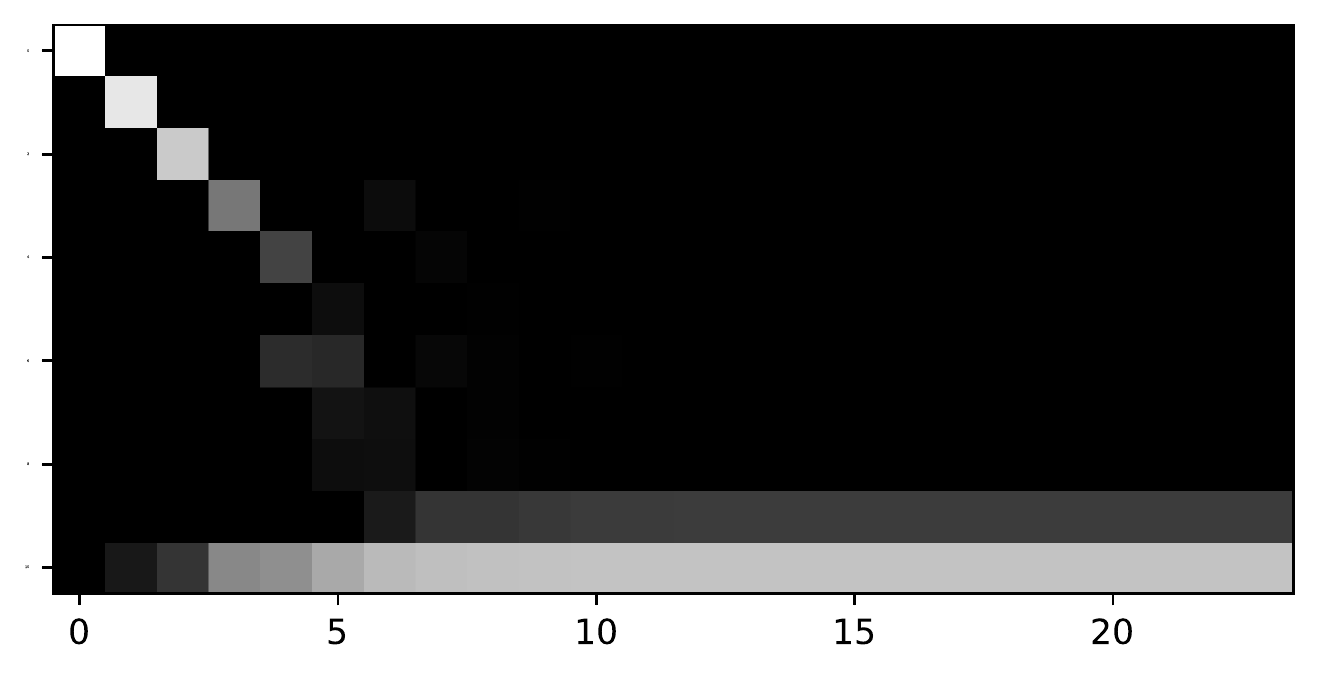}
\caption*{\textsc{Baseline}}
\end{subfigure}
\begin{subfigure}[c]{0.48\textwidth}
\centering
\includegraphics[width=\textwidth]{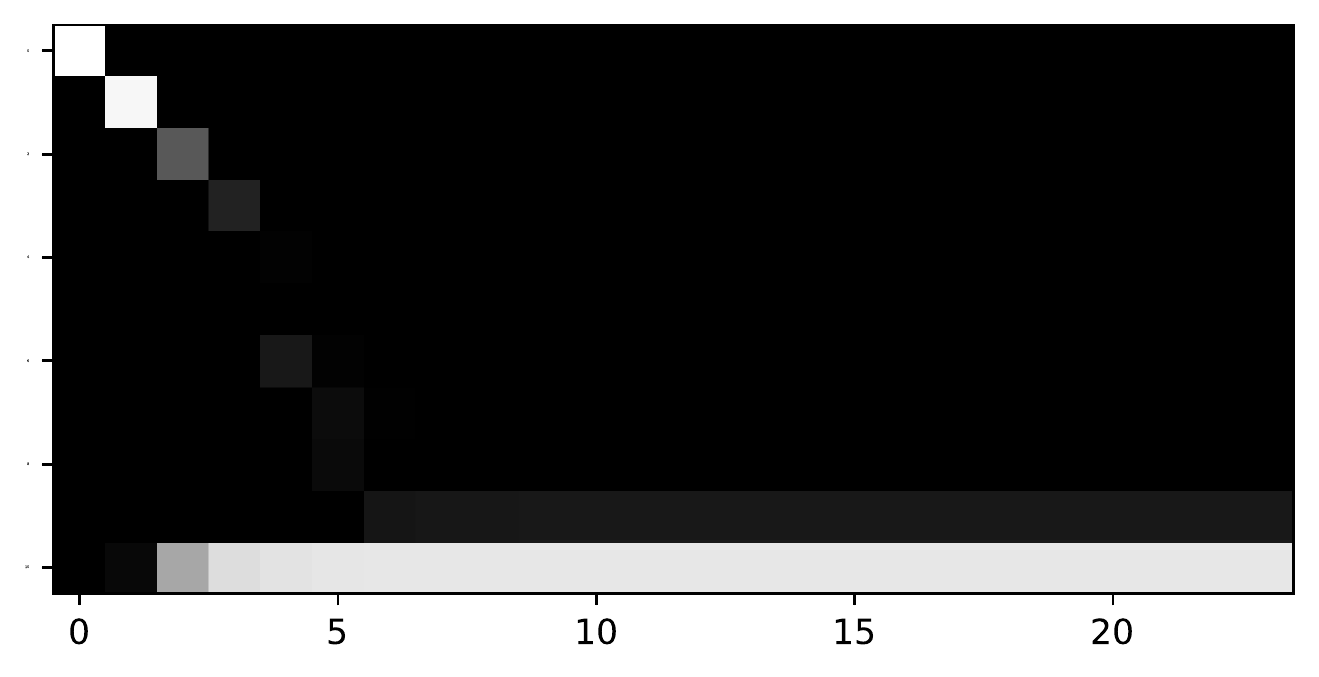}
\caption*{\textsc{Resource Description}}
\end{subfigure}
\end{subfigure}
\end{figure*}
\setcounter{figure}{6}

\begin{figure*}[h]
\caption{
    The target error kind is \textsc{No Error}. \textsc{Baseline} correctly predicts \textsc{No Error} with confidence 0.416. \textsc{Docstring} also correctly predicts \textsc{No Error} with confidence 0.823.
    The \textsc{Baseline} instruction pointer value makes its largest probability mass contribution to $\nerror$ at $n = 0$ and ends up with mass split between $\nexit$ and $\nerror$. The \textsc{Docstring} instruction pointer value accumulates little probability in $\nerror$ and ends up with most probability mass in $\nexit$.
}
\label{tab:visualization-example-p02314-s299863768}

\begin{small}
\begin{subtable}{\textwidth}
\begin{tabularx}{\textwidth}{lX}
\multirow{2}{*}{\textsc{StdIn Description}}
& \code{Input: n m d1 d2 ... dm Two integers n and m are given in the first line. The available denominations are given in the second line.} \\
& \code{Constraints: 1 <= n <= 50000 1 <= m <= 20 1 <= denomination <= 10000 The denominations are all different and contain 1.}
\end{tabularx}
\end{subtable}
\begin{subtable}{\textwidth}
\begin{tabularx}{\textwidth}{cXrr}
\toprule
$n$ & \textsc{Source} & \textsc{Baseline} Error contrib. & \textsc{R.D.} Error contrib. \\
\midrule
\code{0}     & \code{from itertools import combinations\_with\_replacement as C} & 40.1 &  1.3 \\
\code{1}     & \code{n, m = map(int, input().split())}                           &  2.3 &  7.1 \\
\code{2}     & \code{coin = sorted(list(map(int, input().split())))}             &  7.2 &  2.8 \\
\code{3}     & \code{if n in coin:}                                              &  2.0 &  0.2 \\
\code{4}     & \code{~~print(1)}                                                 &  2.0 &  1.5 \\
             & \code{else:}                                                      &      &      \\
\code{5}     & \code{~~end = n // coin[0] + 1}                                   &  0.3 &  0.1 \\
\code{6}     & \code{~~b = False}                                                &  0.1 &  0.3 \\
\code{7,8}   & \code{for i in range(2, end):}                                    &  2.4 &  0.7 \\
\code{9,10}  & \code{~~for tup in list(C(coin, i)):}                             &  3.4 &  1.2 \\
\code{11}    & \code{~~~~if sum(tup) == n:}                                      &  0.3 &  0.0 \\
\code{12}    & \code{~~~~~~print(i)}                                             &  0.3 &  0.1 \\
\code{13}    & \code{~~~~~~b = True}                                             &  0.6 &  0.9 \\
             & \code{~~~~~~break}                                                &      &      \\
\code{14}    & \code{if b:~break}                                                &  0.1 &  1.4 \\
\bottomrule
\end{tabularx}
\end{subtable}
\end{small}

\begin{subfigure}[c]{\textwidth}
\begin{subfigure}[c]{0.48\textwidth}
\centering
\includegraphics[width=\textwidth]{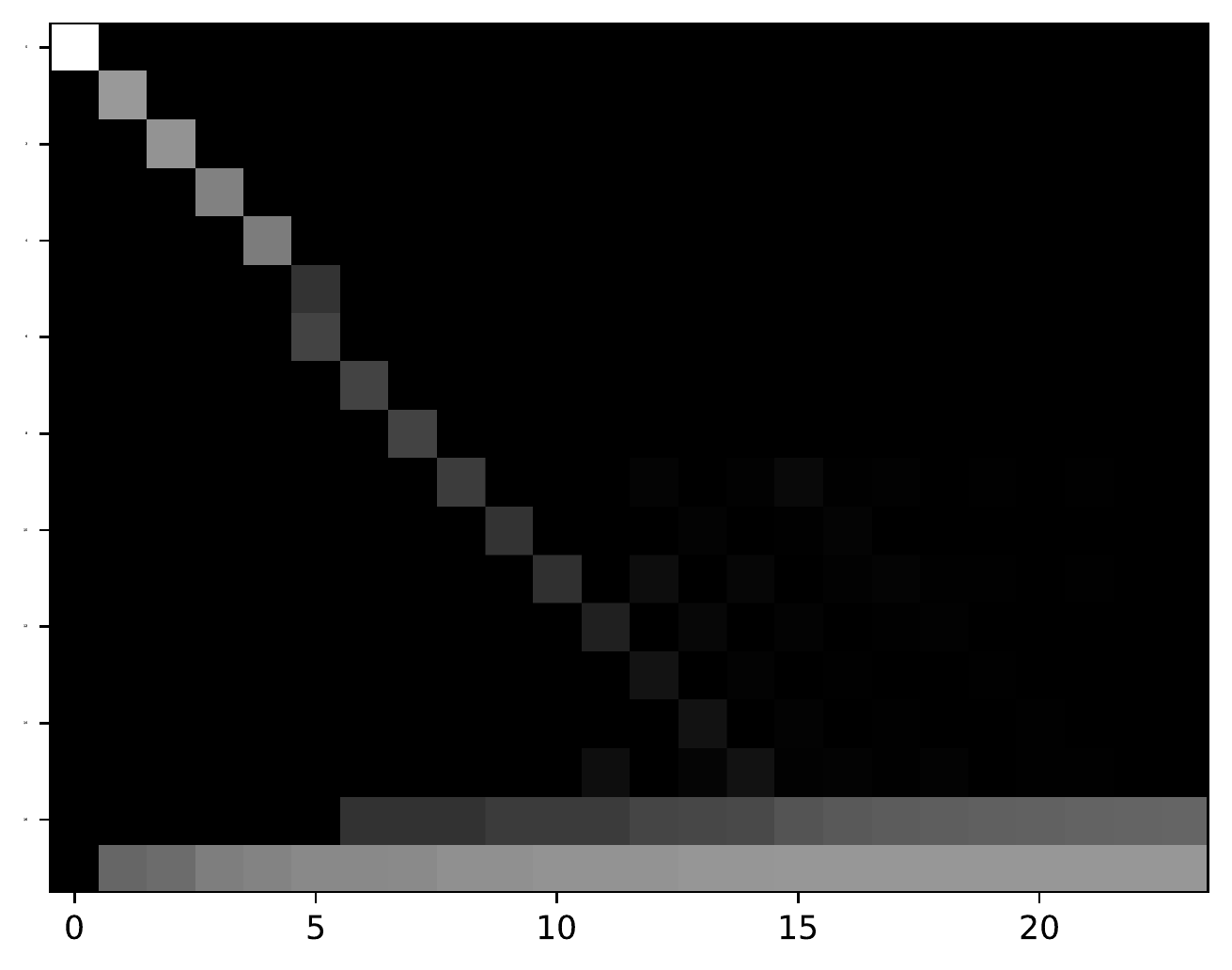}
\caption*{\textsc{Baseline}}
\end{subfigure}
\begin{subfigure}[c]{0.48\textwidth}
\centering
\includegraphics[width=\textwidth]{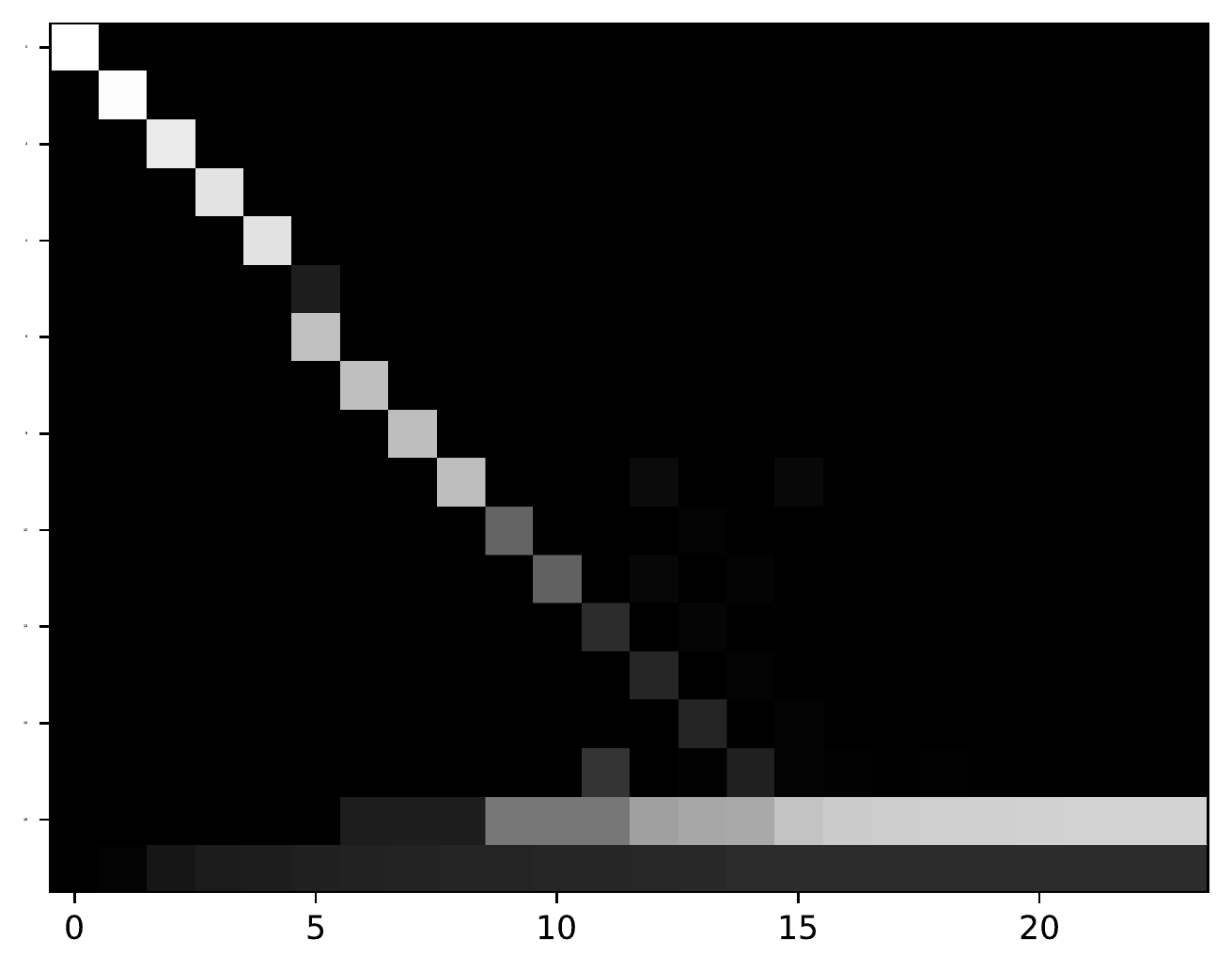}
\caption*{\textsc{Resource Description}}
\end{subfigure}
\end{subfigure}
\end{figure*}

\end{document}